\documentclass[sigconf]{acmart}

\copyrightyear{2025}
\acmYear{2025}
\setcopyright{acmlicensed}
\acmConference[MM '25] {Proceedings of the 33rd ACM International Conference on Multimedia}{October 27--31, 2025}{Dublin, Ireland.}
\acmBooktitle{Proceedings of the 33rd ACM International Conference on Multimedia (MM '25), October 27--31, 2025, Dublin, Ireland}
\acmDOI{10.1145/3746027.3755112}
\acmISBN{979-8-4007-2035-2/2025/10}

\usepackage{epsfig}
\usepackage{algorithm, algpseudocode} %
\usepackage{enumitem}
\usepackage{wrapfig}
\usepackage{pifont} %
\usepackage{stmaryrd} %
\usepackage{booktabs, multirow} %
\usepackage{tablefootnote}
\usepackage{colortbl} %
\usepackage{sidecap} %
\usepackage{dblfloatfix} %
\usepackage{colortbl}
\usepackage{bm}
\usepackage{arydshln}
\usepackage{float}
\usepackage{pgfplots}
\usepackage[utf8]{inputenc}
\usepackage{xspace}
\usepackage[capitalise]{cleveref}
\usepackage{array}       % For m{} column type
\usepackage{multirow}     % For \multirow
\usepackage{balance}
\usepackage{adjustbox} 
\usepackage{arydshln} % Optional: For finer control over scaling/alignment
\usepackage{placeins} % Place this in your preamble
\FloatBarrier
\usepackage{makecell}
\usepackage{capt-of} % Allows captions outside of figure environments
 \usepackage{float}
\pgfplotsset{compat=1.17}
\definecolor{lightblue}{RGB}{202, 240, 248} % Light blue color
\definecolor{lightcream}{RGB}{245, 235, 224}
\definecolor{lightgreen}{RGB}{253, 252, 220}

\usepackage{amsmath}

\usepackage{cleveref}
\Crefname{example}{Example}{Examples}
\usepackage{enumitem}
\newtheoremstyle{abcd}% name
  {}%      Space above, empty = `usual value'
  {}%      Space below
  {\itshape}% Body font
  {}%         Indent amount (empty = no indent, \parindent = para indent)
  {\sc}% Thm head font
  {.}%        Punctuation after thm head
  {.5em}% Space after thm head: \newline = linebreak
  {}%         Thm head spec

\theoremstyle{abcd}

\newtheorem{designchoice}{Design}
\AfterEndEnvironment{designchoice}{\noindent\ignorespaces}

\Crefname{designchoice}{Design}{Design}

\begin{document}

%%
%% The "title" command has an optional parameter,
%% allowing the author to define a "short title" to be used in page headers.
\title{Latent Diffusion Unlearning: Protecting Against Unauthorized Personalization Through Trajectory Shifted Perturbations}

%%
%% The "author" command and its associated commands are used to define
%% the authors and their affiliations.
%% Of note is the shared affiliation of the first two authors, and the
%% "authornote" and "authornotemark" commands
%% used to denote shared contribution to the research.
% \author{Anonymous authors}

\author{Naresh Kumar Devulapally\textsuperscript{\dag}}
\email{devulapa@buffalo.edu}
\thanks{\dag \enspace Corresponding authors: N.~K.~Devulapally and V.~S.~Lokhande.}
\affiliation{%
  \institution{University at Buffalo, The State University of New York}
  \city{Buffalo}
  \state{New York}
  \country{USA}
}

\author{Shruti Agarwal}
\email{shragarw@adobe.com}
\affiliation{%
  \institution{Adobe Research}
  \city{San Jose}
  \state{California}
  \country{USA}
}

\author{Tejas Gokhale}
\email{gokhale@umbc.edu}
\affiliation{%
  \institution{University of Maryland, Baltimore County}
  \city{Baltimore}
  \state{Maryland}
  \country{USA}
}

\author{Vishnu Suresh Lokhande\textsuperscript{\dag}}
\email{vishnulo@buffalo.edu}
\affiliation{%
  \institution{University at Buffalo, The State University of New York}
  \city{Buffalo}
  \state{New York}
  \country{USA}
}

% \author{Lars Th{\o}rv{\"a}ld}
% \affiliation{%
%   \institution{The Th{\o}rv{\"a}ld Group}
%   \city{Hekla}
%   \country{Iceland}}
% \email{larst@affiliation.org}

% \author{Valerie B\'eranger}
% \affiliation{%
%   \institution{Inria Paris-Rocquencourt}
%   \city{Rocquencourt}
%   \country{France}
% }

% \author{Aparna Patel}
% \affiliation{%
%  \institution{Rajiv Gandhi University}
%  \city{Doimukh}
%  \state{Arunachal Pradesh}
%  \country{India}}

% \author{Huifen Chan}
% \affiliation{%
%   \institution{Tsinghua University}
%   \city{Haidian Qu}
%   \state{Beijing Shi}
%   \country{China}}

% \author{Charles Palmer}
% \affiliation{%
%   \institution{Palmer Research Laboratories}
%   \city{San Antonio}
%   \state{Texas}
%   \country{USA}}
% \email{cpalmer@prl.com}

% \author{John Smith}
% \affiliation{%
%   \institution{The Th{\o}rv{\"a}ld Group}
%   \city{Hekla}
%   \country{Iceland}}
% \email{jsmith@affiliation.org}

% \author{Julius P. Kumquat}
% \affiliation{%
%   \institution{The Kumquat Consortium}
%   \city{New York}
%   \country{USA}}
% \email{jpkumquat@consortium.net}

%%
%% By default, the full list of authors will be used in the page
%% headers. Often, this list is too long, and will overlap
%% other information printed in the page headers. This command allows
%% the author to define a more concise list
%% of authors' names for this purpose.
% \renewcommand{\shortauthors}{Trovato et al.}

%%
%% The abstract is a short summary of the work to be presented in the
%% article.
\begin{abstract}
Text-to-image diffusion models have demonstrated remarkable effectiveness in rapid and high-fidelity personalization, even when provided with only a few user images. However, the effectiveness of personalization techniques has lead to concerns regarding data privacy, intellectual property protection, and unauthorized usage. To mitigate such unauthorized usage and model replication, the idea of generating ``unlearnable'' training samples utilizing image poisoning techniques has emerged. Existing methods for this have limited imperceptibility as they operate in the pixel space which results in images with noise and artifacts. In this work, we propose a novel model-based perturbation strategy that operates within the latent space of diffusion models. Our method alternates between denoising and inversion while modifying the starting point of the denoising trajectory: of diffusion models. This trajectory-shifted sampling ensures that the perturbed images maintain high visual fidelity to the original inputs while being resistant to inversion and personalization by downstream generative models. This approach integrates unlearnability into the framework of Latent Diffusion Models (LDMs), enabling a practical and imperceptible defense against unauthorized model adaptation. We validate our approach on four benchmark datasets to demonstrate robustness against state-of-the-art inversion attacks. Results demonstrate that our method achieves significant improvements in imperceptibility ($\sim 8 \% -10\%$ on perceptual metrics including PSNR, SSIM, and FID) and robustness ( $\sim 10\%$ on average across five adversarial settings), highlighting its effectiveness in safeguarding sensitive data. \href{https://github.com/naresh-ub/unlearnable_samples}{\textcolor{blue}{Code}}.
\end{abstract}

%
% The code below is generated by the tool at http://dl.acm.org/ccs.cfm.
% Please copy and paste the code instead of the example below.
%
\begin{CCSXML}
<ccs2012>
   <concept>
       <concept_id>10010147.10010178.10010224.10010240.10010241</concept_id>
       <concept_desc>Computing methodologies~Image representations</concept_desc>
       <concept_significance>500</concept_significance>
       </concept>
 </ccs2012>
\end{CCSXML}

\ccsdesc[500]{Computing methodologies~Image representations}

% \ccsdesc[500]{Do Not Use This Code~Generate the Correct Terms for Your Paper}
% \ccsdesc[300]{Do Not Use This Code~Generate the Correct Terms for Your Paper}
% \ccsdesc{Do Not Use This Code~Generate the Correct Terms for Your Paper}
% \ccsdesc[100]{Do Not Use This Code~Generate the Correct Terms for Your Paper}

%
% Keywords. The author(s) should pick words that accurately describe
% the work being presented. Separate the keywords with commas.
\keywords{Diffusion Models, Unlearning, Personalization, Identity Protection}
%% A "teaser" image appears between the author and affiliation
%% information and the body of the document, and typically spans the
%% page.
% \begin{teaserfigure}
%   \includegraphics[width=\textwidth]{misc/sampleteaser.pdf}
%   \caption{Seattle Mariners at Spring Training, 2010.}
%   \Description{Enjoying the baseball game from the third-base
%   seats. Ichiro Suzuki preparing to bat.}
%   \label{fig:teaser}
% \end{teaserfigure}

% \received{20 February 2007}
% \received[revised]{12 March 2009}
% \received[accepted]{5 June 2009}

%%
%% This command processes the author and affiliation and title
%% information and builds the first part of the formatted document.
\maketitle

\vspace{-0.1in}
\section{Introduction}

Generative AI (GenAI) models, often trained on massive datasets scraped from the web, raise significant ethical and legal concerns particularly around the unauthorized use of copyrighted materials \cite{turnbull2001important}. Given the rapid advancements in model customization or personalization techniques \cite{gal2022imageworthwordpersonalizing, Ruiz_2023_CVPR}, only a few example images are sufficient to closely mimic the style and content of the originals, threatening both individual artists’ community and commercial models \cite{Ruiz_2023_CVPR}. This practice constitutes a clear violation of intellectual property rights and affects negatively to both independent artists and commercial content platforms.

In response to these concerns, several technological solutions have been proposed. One such solution is the use of provenance metadata standards like the C2PA (Coalition for Content Provenance and Authenticity), which includes labels such as “Do Not Train” \cite{C2PA}.  These cryptographic metadata tags are designed to signal creators' preferences regarding the use of their content in AI training pipelines. While this approach has gained traction among some content platforms, its effectiveness is severely limited by the ease with which metadata can be stripped during file handling or sharing. As a result, the creator’s intention may not persist with the digital asset, leading to potential misuse.

Another line of defense involves embedding adversarial perturbations into images to make them “unlearnable”, thereby disrupting the ability of GenAI models to extract meaningful features during training \cite{song2022denoisingdiffusionimplicitmodels, ho2020denoisingdiffusionprobabilisticmodels, Ruiz_2023_CVPR, gal2022imageworthwordpersonalizing}. Existing methods \cite{advdm, simac, metacloak, vcpro, antidb} typically operate in the spatial domain of the image by adding noise patterns that mislead the customization process. However, these techniques suffer from critical drawbacks: the visual quality of the protected image is often compromised, and the perturbations are vulnerable to purification techniques \cite{diffpure} that restore the image’s utility for training.

To address these limitations, we propose a novel method for generating unlearnable examples that are both visually indistinguishable from the original images and robust to purification attacks.
Unlike previous approaches that manipulate pixel-level information, our technique perturbs the latent space, i.e., the internal representations learned by a generative model. This is achieved by learning a latent transformation network that is applied at the beginning of the latent denoising process \cite{nonoise}. The transformed latent, after denoising, remains perceptually similar to the original but cannot be used effectively for downstream personalization tasks.
% This is achieved by learning a latent transformation network that, when applied at the beginning of the latent denoising process, outputs a latent that after denoising remains perceptually similar to the original but cannot be used effectively for downstream personalization tasks such as textual inversion or Dreambooth customization training.

\noindent The contributions and findings of this work are summarized below:
\begin{itemize}[nosep,leftmargin=*]

\item We propose a novel method for generating high-quality unlearnable examples by perturbing the latent space, preventing Textual Inversion (TI) and DreamBooth (DB) based personalization. 

\item Our approach employs a UNet-based architecture to predict unlearnable latent embeddings, eliminating the need for retraining when targeting new concepts. 

\item This approach results in state-of-the-art image quality and robustness against strong purification attacks such as DiffPure. 

\item Results demonstrate that our method generalizes well across different versions of Stable Diffusion and can produce unlearnable examples for identities not seen during training. 

\end{itemize}

\vspace{-0.4in}
\section{Related works}

\paragraph{\textbf{Unlearnable Sample Generation}} Recent efforts leverage adversarial attacks to protect visual content from unauthorized exploitation by generative models. 
Early works primarily focused on discriminative tasks \cite{gan-ian}, but with the rise of diffusion models, researchers now target generative pipelines through image-level perturbations. 
Methods like {Anti-DreamBooth} \cite{antidb} inject adversarial noise to prevent personalization by DreamBooth personalization model. 
Similarly, {Mist} \cite{advdm} proposes techniques to generate adversarial samples that cannot be imitated by Textual Inversion personalization models. Building on meta-learning strategies, {MetaCloak} \cite{metacloak} optimizes perturbations across diverse data transformations to improve robustness against preprocessing attacks. These methods universally apply \textit{global} modifications to the entire image, often resulting in significant noise intensity that degrades visual quality and disproportionately affects the entire image. VCPro \cite{vcpro} points out the lack of imperceptibility for unlearnable sample generation across all the baselines and proposed a region-specific training mechanism for improved imperceptibility while still training on the pixel-space. While imperceptibility is improved, their method requires segmentation masks during training to overlay perturbation on a specific region.

\vspace{-0.4in}

\paragraph{\textbf{Purification Attacks in Latent Diffusion}}{\label{sec:attacks}} In addition to the significant corruption introduced into images in the pixel-space, these methods fail to be robust to many pixel-level transformations \cite{metacloak}. Unlearnable Sample Generation has two critical constraints of imperceptibility and robustness to attacks. In addition to existing image transformation based attacks such as Cropping, JPEG compression, Gaussian Filtering etc., recently strong, adversarial purification method such as DiffPure \cite{diffpure}, WMAttacker \cite{wmattacker} have gained interest. In the presence of such regeneration based adversarial attacks, generating robust, imperceptible adversarial samples becomes increasingly challenging and important to prevent unauthorized use of copyrighted material.
 
\section{Method}

\subsection{Preliminaries}

\paragraph{\textbf{Text-to-Image Diffusion Models}} The diffusion \cite{ho2020denoisingdiffusionprobabilisticmodels} framework consists of two core processes: (1) an \textit{inversion process} $\Phi^{\text{invert}}$ that gradually converts data $x_0$ to noise, and (2) a \textit{denoising process} \(\Phi^{\text{denoise}}\) that removes noise from the noised Gaussian sample \(x_T \sim \mathcal{N}(0, \text{I})\) to recover clean data \(x_0 \in \mathcal{D}\).

\noindent Given a data point $x_0$,the inversion diffusion process $\Phi^{\text{invert}}$, adds noise according to a predefined schedule $\beta_t$:
\begin{equation}
x_T = \Phi^{\text{invert}}_T \circ \Phi^{\text{invert}}_{T-1} \circ \cdots \circ \Phi^{\text{invert}}_1(x_0)
\end{equation}

% where each intermediate step $t$ computes:
% \begin{equation}
% \Phi^{\text{invert}}_t(x_{t-1}) = x_t = \sqrt{1 - \beta_t} \cdot x_{t-1} + \sqrt{\beta_t} \cdot \epsilon_t, \quad \epsilon_t \sim \mathcal{N}(0, I)
% \end{equation}

\noindent Conversion of \( x_0 \) to \( x_t \) using cumulative noise scaling \( \bar{\alpha}_t = \prod_{i=1}^t (1 - \beta_i) \) in one-step is given by:
\begin{equation}
    x_t = \sqrt{\bar{\alpha}_t} \, x_0 + \sqrt{1 - \bar{\alpha}_t} \, \epsilon, \quad \epsilon \sim \mathcal{N}(0, I)
    \label{eq:reparam}
\end{equation}
As \( T \to \infty \), \( x_T \) converges to standard Gaussian noise \( \mathcal{N}(0, I) \). The schedule \( \beta_t \) controls the rate of corruption, ensuring the reverse process \( \Phi^{\text{denoise}} \) can learn to iteratively denoise the data \cite{song2022denoisingdiffusionimplicitmodels}.

\noindent Given a text prompt $p$ and guidance scale \cite{cfg} \(\gamma_1\), the text-conditioned denoising process operates over \(T\) steps, decomposed as:
\begin{align}
    \Phi^{\text{denoise}}(x_T|p, \gamma_1) 
    = \resizebox{.25\textwidth}{!}{$
    \Phi^{\text{denoise}}_0 \circ \Phi^{\text{denoise}}_{1} \circ \cdots \circ \Phi^{\text{denoise}}_T(x_T|p, \gamma_1)
    $}
\end{align}
Thus the reconstructed image is given by $\bar{x}_0 = \Phi^{\text{denoise}}(x_T|p, \gamma_1)$ and each step \(t\) refines the data using:
\begin{equation}
    \bar{x}_{t-1} = \sqrt{\alpha_{t-1}} \frac{\bar{x}_t - \sqrt{1 - \alpha_t}\epsilon_\theta^t(\bar{x}_t)}{\sqrt{\alpha_t}} + \sqrt{1 - \alpha_{t-1}}\epsilon_\theta^t(\bar{x}_t).
\end{equation}
Here, \(\alpha_t = (1 - \beta_t)\) follows a predefined noise schedule \(\beta_t\) \cite{song2022denoisingdiffusionimplicitmodels}, and \(\epsilon_\theta^t\) is a neural network predicting noise at step \(t\). 
Under the approximation \(\epsilon^t(x_t) \approx \epsilon^t(x_{t-1})\) \cite{nulltext}, \(\Phi^{\text{denoise}}\) becomes the exact inverse of the forward process \(\Phi^{\text{invert}}\), enabling bidirectional mapping between noise and data. This property enables faithful reconstruction or controlled editing of clean data via diffusion steps. With the forward and reverse processes defined, the model is trained by minimizing a loss that predicts the noise at each time step. 
\begin{equation}
    \mathcal{L}_{\text{DM}} = \; \mathbb{E}_{\substack{x,y,\epsilon \sim \mathcal{N}(0,1),t}} 
\Bigl[
\|\epsilon - \epsilon_\theta\bigl(x_t,\, t,\, \mathcal{T}(y)\bigr)\|_2^2
\Bigr]
\label{eq:ldm_loss}
\end{equation}

\noindent where $\mathcal{T}$ is the Text Encoder and Diffusion Model trains parameters $\epsilon_\theta(\cdot)$.

% \subsection{Preliminaries on Latent Diffusion Models}

% \begin{itemize}
%     \item Latent Diffusion Forward Process Equation
%     \item Noise matching loss and Denoising Equations
%     \item Classifier-free guidance
%     \item Distribution information $p(z \mid x)$ at final noise timestep $T$.
% \end{itemize}
\subsubsection{Diffusion Models for Personalized Content Generation} Textual Inversion (TI) \cite{gal2022imageworthwordpersonalizing} is a personalization technique that introduces a new pseudo-token $S_*$ into the text encoder of an LDM pipeline and then fine-tunes the token's embeddings for personalization. To fine-tune the token embeddings of $S_*$, TI uses direct optimization by minimizing the LDM loss by conditioning the denoising process in the presence of $S_*$ embeddings. The personalization loss objective is given by:

\vspace{-0.2in}

\begin{equation}
    \mathcal{L}^{\text{TI}}_{\text{personalize}} = \; \mathbb{E}_{\substack{x,\epsilon \sim \mathcal{N}(0,1),t}} 
\Bigl[
\|\epsilon - \epsilon_\theta\bigl(x_t,\, t,\, \mathcal{T}_{\theta}(S_*)\bigr)\|_2^2
\Bigr]
\label{TI_loss}
\end{equation}

% \begin{figure*}[!t]
%     \centering
%     \includegraphics[width=\linewidth]{figures/training_updated.jpeg}
%     \vspace{-6mm}
%     \caption{Training Pipeline: \textmd{\textit{We train the parameters of unlearnable pertubation model `$\rho$' that perturbs the initial point of the denoising trajectory in the latent space to generate $\bar{z}^{\text{ul}}_T$ followed by shortcut diffusion model with (k=4). This is followed by a pre-trained, frozen personalization model ($\Phi^{\text{personalize}}$) to maximize $\mathcal{L}_{\text{personalize}}$. }}}
%     \label{fig:training_pipeline}
% \end{figure*}

\begingroup
% tighten space below two-column floats and inside caption
\setlength{\dbltextfloatsep}{4pt plus 1pt minus 1pt}
\setlength{\textfloatsep}{4pt plus 1pt minus 1pt}
\captionsetup[figure]{aboveskip=2pt, belowskip=-10pt}

\begin{figure*}[!t]
    \centering
    \includegraphics[width=\linewidth]{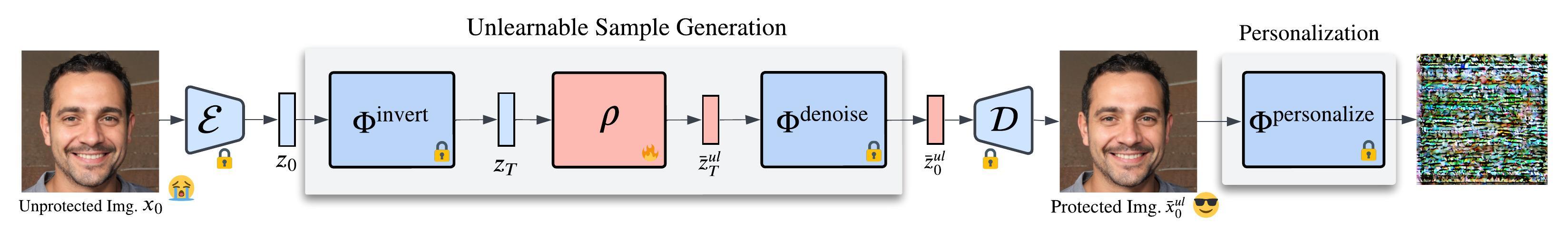}
    \caption{Training Pipeline: \textmd{\textit{We train the parameters of unlearnable pertubation model `$\rho$' that perturbs the initial point of the denoising trajectory in the latent space to generate $\bar{z}^{\text{ul}}_T$ followed by shortcut diffusion model with (k=4). This is followed by a pre-trained, frozen personalization model ($\Phi^{\text{personalize}}$) to maximize $\mathcal{L}_{\text{personalize}}$. }}}
    \label{fig:training_pipeline}
\end{figure*}
\endgroup
% \vspace{-4pt}

\noindent where $\mathcal{T}_\theta$ represents the Text Encoder and the embeddings of $S_*$ are fine-tuned within $\mathcal{T}_\theta$, $z_t$ represents noised latent at timestep $t$, $\epsilon$ is the noise added during the Forward Diffusion Process. Utilizing this formulation TI is capable of performing personalization with a small set of input images (3-5). The optimal token embeddings for are obtained by minimizing $\mathcal{L}^{\text{TI}}_{\text{personalize}}$.

DreamBooth is another popular personalization technique that targets minimizing the personalization loss given by:

\begin{equation}
    \mathcal{L}^{\text{DB}}_{\text{personalize}} = \mathbb{E}_{\substack{x_p, x_{\text{cls}}, \epsilon, t}} 
    \resizebox{0.25\textwidth}{!}{$\left[
    \left\| \epsilon - \epsilon_\theta(x_p^t, t, c_p) \right\|^2 + \lambda \left\| \epsilon - \epsilon_\theta(x_{\text{cls}}^t, t, c_{\text{cls}}) \right\|^2
    \right]$}
\end{equation}

\noindent where $x_p$: Few-shot subject images, $x_{\text{cls}}$: Class images;  
$c_p$: Unique-text prompt, $c_{\text{cls}}$: Class-text prompt;  
$\lambda$: Prior preservation weight.

\noindent From here on, we use $\mathcal{L}_{\text{personalize}}$ as $\mathcal{L}^{\text{TI}}_{\text{personalize}}$ by default.

\subsection{Objective for Diffusion Unlearning}
Given an original image $x_0$, our goal is to construct an unlearnable variant of this image, denoted by $\bar{x}_0^{\text{ul}}$. The defining characteristic of $\bar{x}_0^{\text{ul}}$ is that it should effectively disrupt a downstream personalization pipeline, such as Textual Inversion. In other words, $\bar{x}_0^{\text{ul}}$ should render these pipelines incapable of extracting or encoding meaningful representations of the original subject depicted in $x_0$. Importantly, this unlearnability objective must remain robust even in the presence of purification defenses applied post hoc to $\bar{x}_0^{\text{ul}}$. As discussed in the previous section, several purification strategies exist, with DiffPure \cite{diffpure} representing one of the most effective. The design of $\bar{x}_0^{\text{ul}}$ must, therefore account for such countermeasures, ensuring that the adversarial perturbation cannot be reversed or neutralized through purification. In the following subsections, we discuss the core objectives that guide the construction of $\bar{x}_0^{\text{ul}}$.

\vspace{-0.05in}

\paragraph{\textbf{Obstructing Personalized Generation}} We define an unlearnable image parameterized by $\rho$, denoted as $\bar{x}_0^{\text{ul}}(\rho)$, where $\rho$ controls the generative process. Details regarding the nature and optimization of $\rho$ will be discussed in subsequent sections. For now, we assume a parameterized pipeline exists to synthesize $\bar{x}_0^{\text{ul}}(\rho)$. The goal is to disrupt the effectiveness of personalization pipelines. This can be formulated as the maximization of the personalization loss $\mathcal{L}_{\text{personalize}}$ applied to the reconstructed output

\vspace{-0.15in}

\begin{align}
    \max_{\rho} \mathcal{L}_{\text{personalize}} \left( \bar{x}_0^{\text{ul}}(\rho) \right)
\end{align}
%where $\mathcal{A}$ denotes a purification method aimed at removing adversarial perturbations. 
Next, we describe how $\bar{x}_0^{\text{ul}}(\rho)$ is generated using image-to-image translation within an LDM framework.

\vspace{-0.05in}

\paragraph{\textbf{Generating Unlearnable Images through Penalty-based Optimization}} We now shift from pixel-level (image-level) diffusion models to Latent Diffusion Models. Let $\mathcal{E}$ be an encoder that converts an image $x_0$ to latent $z_0$, and $\mathcal{D}$ be a decoder that reconstructs $x_0$ to latent $z_0$.

Given a reference image $x_0$, we obtain the latent code $z_0$ and then our method generates an unlearnable latent $\bar{z}^{\text{ul}}_0(\rho)$ as shown in \cref{fig:training_pipeline}. This unlearnable latent $\bar{z}^{\text{ul}}_0(\rho)$ is optimized to ensure that the imperceivable budget constraint for unlearning $\delta^u$ is met at the pixel-level $\bar{x}^{\text{ul}}_0$.

To train for $\bar{z}^{\text{ul}}_0(\rho)$ generation we utilize a pre-trained personalization model and generate $\bar{z}^{\text{ul}}_0(\rho)$ to maximize $\mathcal{L}_{\text{personalize}}$. Our learning objective is given by:

\begin{equation}
\begin{aligned}
    &\text{maximize } \mathcal{L}_{\text{personalize}}(\bar{z}^{\text{ul}}_0) \\
    &\text{such that } || (\bar{x}^{\text{ul}}_0) - (x_0) || \leq \delta^u
\end{aligned}
\label{eq:overall_loss}
\end{equation}

where personalization loss at the latent-level is given by:
\begin{equation}
    \mathcal{L}_{\text{personalize}}(\bar{z}^{\text{ul}}) = \; \mathbb{E}_{\bar{z}^{\text{ul}}, \substack{t}} 
\Bigl[
\|\epsilon - \epsilon_\theta\bigl(\bar{z}^{\text{ul}}_t,\, t,\, \mathcal{T}(y)\bigr)\|_2^2 \Bigr]
\end{equation}

\noindent We utilize Lagrangian Multiplier-based solution \cite{vcpro} to jointly optimize for imperceptibility constraint and unlearning objective that achieves the unlearning objective while being imperceptibly perturbed.
\begin{equation}
    \mathcal{L} = \lambda \cdot (|| (\bar{x}^{\text{ul}}_0) - (x_0) || - \delta^u) + (-\mathcal{L}_{\text{personalize}}( \bar{z}^{\text{ul}}_0)),
\end{equation}
where $\delta^u$ is the perturbation budget in the image-space, $\lambda$ is a hyper-parameter. We follow \cite{bertsekas2014constrained} to find $\lambda$ by linearly scheduling $\lambda$ to increase until the imperceptibility constraint is achieved.

% we generate its unlearnable counterpart via image-to-image translation using a latent diffusion model. The diffusion model loss, denoted by $\mathcal{L}_{\text{DM}}$, has been introduced in the preliminaries. Here, we utilize $\mathcal{L}_{\text{DM}}$ to optimize the parameters $\theta$ that define the unlearnable image $x_0^{\text{ul}}(\theta)$, while enforcing a constraint that the generated image remains visually similar to the original $x_0$. This similarity constraint is imposed by a bounded $\ell_2$ norm, parameterized by a deviation budget $\delta$, ensuring that perturbations remain imperceptible or minimally detectable. The optimization objective can be formulated as:
% \begin{align}
%     \min_{\theta} \mathcal{L}_\text{DM} \left(x_0^{\text{ul}}(\theta), x_0\right) \\
%     \text{s.t.} \quad \|x_0^{\text{ul}}(\theta) - x_0\|_2 \le \delta
% \end{align}
% This formulation ensures that the poisoned image $x_0^{\text{ul}}$ remains close to $x_0$ in pixel space while being adversarially crafted to interfere with downstream personalization. In the following section, we detail how latent-space perturbations are used to jointly optimize this constraint alongside the unlearning objective.
\begingroup
% tighten space below two-column floats and inside caption
\setlength{\dbltextfloatsep}{4pt plus 1pt minus 1pt}
\setlength{\textfloatsep}{4pt plus 1pt minus 1pt}
\captionsetup[figure]{aboveskip=2pt, belowskip=-10pt}
\begin{figure*}[!t]
    \centering
    \includegraphics[width=\linewidth]{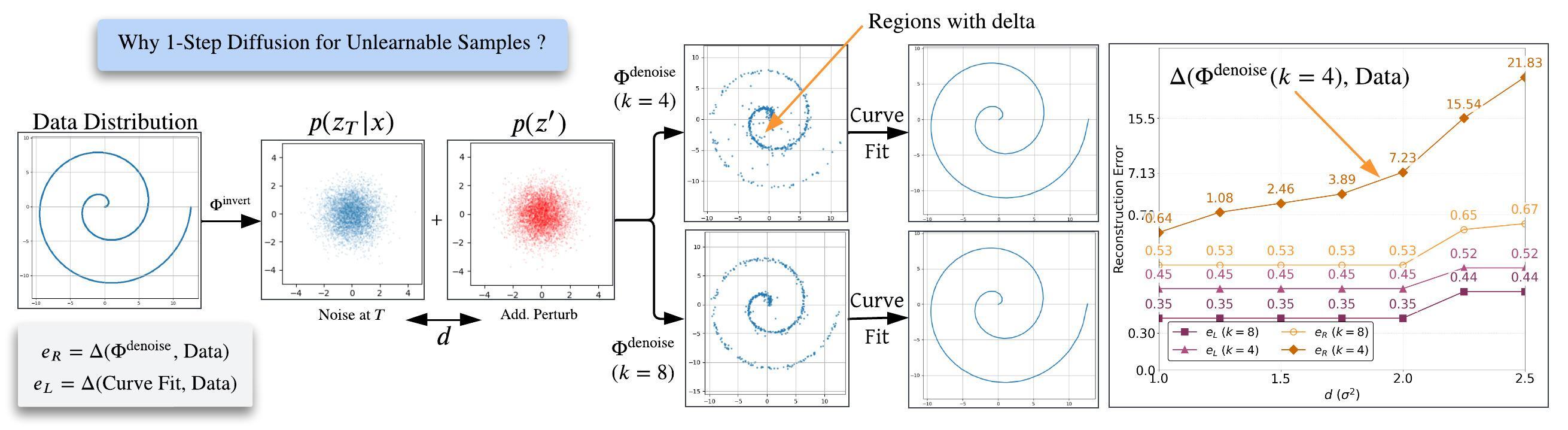}
    % \vspace{-6mm}
    \caption{Few-step Diffusion Models maintain data distribution integrity while allowing perturbations in reconstruction: \textmd{\textit{Our method performs unlearnable sample generation by perturbing the noised latent $z_T$ followed a denoising model $\Phi^{\text{denoise}}$ to generate $\bar{z}^{\text{ul}}_0$. We empirically analyze the properties of $\Phi^{\text{denoise}}$ that allow meaningful perturbations $\Delta z_T$ to survive while preserving the underlying data distribution. Using a spiral dataset, we compare curve-fit error $e_L = \Delta(\text{Curve Fit}, \text{Data})$ and sample-level reconstruction error $e_R = \Delta(\Phi^{\text{denoise}}, \text{Data})$ across different denoising step counts $k$. Results show that multi-step denoising ($k=8$) minimizes both $e_L$ and $e_R$, suppressing perturbations and thus limiting adversarial behavior. In contrast, fewer denoising steps ($k=4$) enable high $e_R$ with low $e_L$, identifying a ``feasible region” where latent perturbations persist through $\Phi^{\text{denoise}}$ while maintaining distributional integrity—critical for imperceptible unlearnable signal injection. This shows few-step (particularly 1-step) diffusion models as optimal for latent-space unlearnable sample generation. We provide the trends on $e_R$ and $e_L$ at different $k$ values in the plot (Right). \textbf{Curve-Fit here is used to assess if distribution integrity is maintained.}}}}
    \label{fig:method_motivation}
\end{figure*}
\endgroup
\vspace{-6pt}

% \begin{figure}[!t]
%     \centering
%     \includegraphics[width=0.6\linewidth]{figures/side_fig.jpeg}
%     \caption{Caption: \textmd{\textit{Update caption here.}}}
%     \label{fig:budget_qual}
% \end{figure}

\subsection{Altering the Diffusion Trajectory through Initial Latent Perturbation}{\label{sec:altering}} In this section we introduce a critical design choise of our method which performs latent perturbation at the final timestep $T$ of the Forward Diffusion Process. Consider an inversion operation on latent code $z_0$, defined by $\Phi^\text{invert}$. The inversion yields a noisy latent code, $z_T = \Phi^\text{invert}(z_0)$ (observe that $\Phi^\text{invert}$ has no condition w.r.t. prompt). Subsequently, we define a denoising operator $\Phi^\text{denoise}$, that reconstructs the latent code from the noisy latent state  $z_T$ conditioned on $p$, given by $\bar z_0 =  \Phi^\text{denoise}(z_T \mid p)$. Here,  $\bar z_0$ denotes the reconstructed version of the original latent representation $z_0$. This formulation provides a foundation for manipulating and optimizing trajectories within the latent space while maintaining semantic consistency through the inversion–denoising loop.
% \begin{wrapfigure}{r}{0.4\linewidth}
%     \centering
%     \includegraphics[width=0.58\linewidth]{figures/side_fig.jpeg}
%     \caption{Caption: \textmd{\textit{Update caption here.}}}
%     \label{fig:budget_qual}
% \end{wrapfigure}

While the inversion operator $\Phi^{\text{invert}}$ remains fixed to ensure consistency in mapping from the latent representation to the noisy latent code, the denoising operator $\Phi^{\text{denoise}}$ can be varied to yield diverse reconstructions.  Modifying the denoising process effectively corresponds to adopting an alternative sampling strategy, which can produce significantly different reconstructed outputs. Prior work \cite{nonoise, yoon2024safree, ma2025inferencetimescalingdiffusionmodels, liu2024alignmentdiffusionmodelsfundamentals} suggest that the variability in reconstructed images can be efficiently induced by perturbing the initial noise vector—equivalently, the terminal latent state $z_T$—rather than modifying the entire denoising trajectory across $T$ timesteps. This insight motivates the focus on targeted perturbation of $z_T$ as a computationally efficient and controllable mechanism for altering the final reconstruction. 

\begin{wrapfigure}{r}{0.45\linewidth}
  % Reduce padding locally
  % \setlength{\intextsep}{0pt}%
  % \vspace{-1\baselineskip}  % Pull figure up, if desired
  \centering
  \vspace{-1.2\intextsep}
  \includegraphics[width=0.86\linewidth]{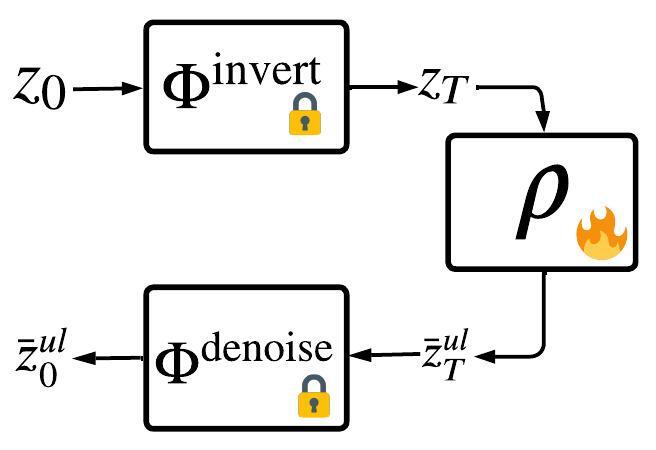}
  \caption{\textmd{\textit{$\rho$ is trained in the presence of $\Phi^{\text{denoise}}$ to get $\bar{z}^{\text{ul}}_T$.}}}
  \label{fig:budget_qual}
  \vspace{-\intextsep}
  % \vspace{-1\baselineskip} % Reduce space after figure
  \vspace{0.1in}
\end{wrapfigure}

The central question then becomes: \textit{how can we perturb $z_T$ in a principled manner to produce poisoned reconstructions that render the image unlearnable for downstream personalization tasks?} A variety of transformation functions $\rho: \mathbb{R}^d \rightarrow \mathbb{R}^d$ can be employed for this purpose, where $d$ is the dimensionality of $z_T$. These transformations range from simple affine maps (e.g., scale-shift) to expressive parameterizations such as fully connected neural networks.

In this work, we adopt a single-layer U-Net architecture \cite{peng2024unetv2rethinkingskip} to parameterize $\rho$, leveraging its ability to produce structured, spatially-aware perturbations informed by contextual information. 
Skip connections within U-Net architectures tend to constrain output variance, promoting stability while still enabling meaningful transformation of $z_T$ \cite{peng2024unetv2rethinkingskip, ronneberger2015unetconvolutionalnetworksbiomedical}. This leads us to the following sampling strategy:

% While the inversion process $\Phi^\text{invert}$ will remain intact, the denoising process $\Phi^\text{denoise}$ can be allowed to vary, to generate different reconstructed images. A different denoising process suggests an improved sampling method that can generate very different reconstruted images. The question then comes to what new denoising process to choose. Research suggests that controlling the initial noise (an equivalent of $z_T$) is sufficient in generating very different reconstructed images. Perturbing initial noize $z_T$ is very much more efficient than perturbing all of the steps of a $T$ step denoising processes. The next question then is how to controllably perturb $z_T$ so as to poison the recontructed image for unlearnability of the downstream task. Various transformations exist $\rho: \bbR^d \rightarrow \bbR^d$ for a $d-$ dimensional $z_T$, ranging from fully-connected networks to scale-shift transformations. In this work, we will resort to one layer of UNET \cite{}. UNET offeres structured, context-aware perturbations that can be introduced on top of $z_T$ \cite{}. Moreover, the skip connections do not permit significant variance in the outputs. Thus we make the following design choice, 

\vspace{-0.03in}

\begin{designchoice}
    {\bf Sampling Strategy for Latent Diffusion Unlearning}. Given an input image $x_0$ or its corresponding latent representation $z_0$, we generate an unlearnable image $\bar{x}_0^{\text{ul}}$ (or its latent counterpart $\bar{z}_0^{\text{ul}}$) via the following sequence of operations:
    \begin{enumerate}[leftmargin=*,nosep]
        \item \textbf{Latent Inversion:} Apply the inversion process to obtain terminal latent state $z_T = \Phi^{\text{invert}}(z_0)$.
        \item \textbf{Perturbation:} Introduce a structured perturbation to the inverted latent using a parameterized transformation $\rho$, implemented as a U-Net module: $\bar{z}_T^{\text{ul}} = \rho(z_T) := \text{UNET}(z_T)$.
        \item  \textbf{Denoising:} Perform the denoising process to reconstruct the perturbed latent: $\bar{z}_0^{\text{ul}} = \Phi^{\text{denoise}}(\bar{z}_T^{\text{ul}} \mid c)$.
    \end{enumerate}
\end{designchoice}

\vspace{-2mm}

Notably, if the transformation $\rho$ were the identity function, i.e., $\rho(z_T) = z_T$, then the output would reduce to the original reconstruction: $\bar{z}_0^{\text{ul}} = \bar{z}_0$.

\vspace{-2mm}

\subsection{Few-Step Denoising for Reliable Semantic Preservation}

In the preceding section, we introduced our design choice to perturb the noisy latent representation $z_T$ with the goal of generating unlearnable latent $\bar{z}^{\text{ul}}_0$. However, for the perturbed output $\bar{z}_0^{\text{ul}}$ to remain effective in evading model personalization while still appearing visually consistent with the original input $z_0$, it is essential to ensure that semantic fidelity is preserved during the reconstruction process. At this point, we encounter a trade-off between the trainable U-Net $\rho$ that aims to push $\bar{z}_0^{\text{ul}}$ away from $z_0$ to bring unlearnable behavior, while the denoising U-Net $\Phi^{\text{denoise}}$ is trained to reconstruct $z_0$ from a noised $z_T$. To study this trade-off we perform an empirical analysis on the \textit{desired behaviour of $\Phi^{\text{denoise}}$ for imperceptible unlearnable sample generation}.

% The standard denoising procedure $\Phi^{\text{denoise}}$ typically consists of $T$ discrete steps, each approximating a continuous reverse-time stochastic differential equation. Formally, the full denoising process can be expressed as a composition of $T$ sequential operations: 
% %
% \begin{align} 
% \Phi^{\text{denoise}}(z_T | p) = \Phi^{\text{denoise}}_{T} \circ \Phi^{\text{denoise}}_{T-1} \circ \cdots \circ \Phi^{\text{denoise}}_{0}(z_T \mid p) 
% \end{align}
% %
% where $p$ denotes the conditioning signal, e.g., textual prompt or class embedding. In the idealized case where the denoising process is governed by a perfectly learned continuous-time ODE, the mapping from noise to data would be deterministic and stable. However, this assumption breaks down under discretized schedules, particularly when using a finite number of steps. Moreover, multi-step denoising is computationally intensive and time-consuming, making it less desirable for scalable applications.

\textbf{\textit{Example 1:}} Our empirical observations are illustrated in \cref{fig:method_motivation}. To perform this study, we consider a simple setting to reconstruct a spiral data distribution. We apply a \textit{random} perturbation to the latent $z_T$ and study the reconstruction using $\Phi^{\text{denoise}}$. We then perform Curve-Fit using a Neural Network to validate if the data distribution is maintained after denoising perturbed $z_T$. Our experiments reveal the trade-off between $\rho$ and $\Phi^{\text{denoise}}$. 

As demonstrated in Figure~\ref{fig:method_motivation}, multi-step denoising processes exhibit strong convergence properties that actively suppress latent-space perturbations during reconstruction—preserving data distribution reconstruction by reducing the reconstruction error ($e_L$) but limiting adversarial effectiveness through reduced reconstruction error ($e_R$). Conversely, we identify that \textit{fewer denoising steps create a ``feasible region"} where significant latent perturbations can be embedded while maintaining distributional integrity ($e_L$) and perceptual similarity. This phenomenon arises because denoising trajectories lack the iterative refinement needed to fully reverse structured perturbations, enabling persistent adversarial signals. We study this behaviour using Single-step denoising models \cite{onestep} that prove particularly advantageous allowing targeted perturbations to propagate directly to the output, achieving the dual objectives of (1) preserving imperceptibility by maintaining distributional integrity (imperceptibility) while (2) Providing
``feasible region" for latent-level perturbation (desirable to bring unlearnable behaviour). Thus, we identify that 1-Step Diffusion models serve our purpose in the best way to perform imperceptible (in the image-level) perturbations at the latent-space for unlearnable sample generation. We identify the optimal steps for $\Phi^{\text{denoise}}$ to be 4 $(k=4)$ and present our results. Few-step diffusion updates, with step-size $d$ are given by:

\vspace{-5mm}

\begin{align}
    \resizebox{.42\textwidth}{!}{$
    \Phi^{\text{denoise}}(z_T \mid p, d) := \Phi^{\text{denoise}}_d(z_d, d) \circ .. \circ \Phi^{\text{denoise}}_{T-d}(z_{T-d}, d) \circ \Phi^{\text{denoise}}_T(z_T, d)
    $}
    \label{eq:reverse_process}
\end{align}

\newcommand{\bignum}[1]{\Large #1}
\section{Experiments}

\subsection{Experimental Setup}

\paragraph{Datasets} We evaluate our method on four datasets, namely, CelebA-HQ \cite{celeba}, VGGFace2 \cite{vggface2}, WikiArt, and DreamBooth datasets. Within these four datasets, we categorize Celeba-HQ, VGGFace2 as Face datasets, and WikiArt \cite{wikiart}, DreamBooth \cite{Ruiz_2023_CVPR} as Non-Face datasets. During training, (for Face datasets) we utilize the dataset provided in \cite{antidb} (Celeba-HQ and VGGFace2 subsets). We utilize 50 identities from each dataset containing a subset of 6 images for each individual during training. For Non-Face datasets, we utilize 30 objects provided in DreamBooth dataset (each with $\sim$ 5 images) and 15 art styles from WikiArt dataset (each with $\sim$ 6 images).

\subsection{Evaluation Metrics} 

Recall from \cref{sec:altering} that our method aims to alter the Diffusion Trajectory to generate imperceptible unlearnable images. 
For a given image $x_0$, (latent code $z_0$), we generate the unlearnable sample $\bar{x}^{\text{ul}}_0$ (latent code $\bar{z}^{\text{ul}}_0$) that cannot be utilized for personalization.
We use three evaluation metrics for comparisons:

\paragraph{\textbf{Imperceptibility}} We utilize Peak Signal-to-Noise Ratio (PSNR) \cite{psnr}, Structural Similarity Index (SSIM) \citep{ssim, nilsson2020understandingssim}, and Fréchet inception distance (FID)~\cite{jayasumana2024rethinkingfidbetterevaluation} to evaluate imperceptibility between original image $x_0$ and the unlearnable image $\bar{x}^{\text{ul}}_0$, which is an important measure for this task.
% This is an important constraint that the unlearnable perturbation is imperceptible to the naked eye. 
% This is ensured by validating imperceptibility metrics 

\paragraph{\textbf{Countering Personalization after $\bar{x}^{\text{ul}}_0$ generation}} Once a subset of unlearnable samples $\{\bar{x}^{\text{ul}}_0\}$ are generated they are sent into a personalization pipeline. We utilize Textual Inversion (TI) for personalization. The aim of unlearnable samples $\{\bar{x}^{\text{ul}}_0\}$ is to perform poorly on personalization metrics. We evaluate performance of countering personalization using metrics including Identity Score Matching (ISM) \cite{9157330}, Blind/Referenceless Image Spatial Quality Evaluator (BRISQUE) \cite{brisque}, Face Detection Failure Rate (FDFR) \cite{fdfr} for Face datasets, and Subject Detection Score (SDS) \cite{9157330}, Identity Score Matching (ISM), Blind/Referenceless Image Spatial Quality Evaluator (BRISQUE), we utilize CLIP-IQAC \cite{Wang_Chan_Loy_2023}, and LIQE \cite{brisque} from \cite{metacloak} for Non-Face datasets. 

\paragraph{\textbf{Robustness of $\bar{x}^{\text{ul}}_0$}} Attack module is placed in between Unlearnable Sample Generation (creation of $\bar{x}^{\text{ul}}_0$) and Personalization using $\bar{x}^{\text{ul}}_0$. Our method is validated to check robustness to basic and adversarial based attack modules. Recall \cref{sec:attacks} that introduces DiffPure (a strong reconstruction-based attacks that utilizes Diffusion Denoising). Robustness to such strong adversarial attacks motivates our method to perform latent-level perturbation for unlearnable sample generation. We identify that existing pixel-level perturbation based method are not robust to denoising based purification attacks such as DiffPure. We emphasize that a robust unlearnable sample generation method is expected to be robust to denoising based purification attacks due to growing prevalence of such attacks \cite{wmattacker, sdedit, diffpure}. Hence, we evaluate and present our results in the presence of DiffPure attack.

\vspace{-0.1in}

\subsection{Training Configuration}{\label{sec:training-conf}}

Our method utilizes Latent Diffusion based reconstruction pipeline for training. The Stable Diffusion (SD) v2-1-base \citep{Rombach_2022_CVPR} is used as the model backbone by default. We utilize a subset from CelebA-HQ and VGGFace2 dataset, grouped by identities, for training, to generate unlearnable samples with minimal perturbations that close to input images. For Textual Inversion training, we fine-tune the text-encoder for an ``sks'' token with a learning rate of $5 \times 10^{-4}$ with $1500$ max training steps with the batch size of $1$. For DreamBooth training, we follow \cite{metacloak}, to fine-tune the U-Net model with a learning rate of $5 \times 10^{-7}$ and batch size of 2 for 1000 iterations in mixed-precision training mode.

\paragraph{\textbf{Baselines Implementation Details}} We compare our method with a) AntiDB \cite{antidb}: that proposes Fully-trained Surrogate Model Guidance (FSMG) and Alternating Surrogate and Perturbation Learning (ASPL) that learn adversarial noise to counter DreamBooth personalization.  b) Mist (AdvDM) \cite{advdm}: that leverages a pre-trained diffusion model for generating adversarial examples to counter Textual Inversion. c) AntiDB $+$ SimAC \cite{simac}: that proposes a greedy timestep selection in the forward diffusion process followed by reverse diffusion to learn unlearnable perturbation in the image space. d) Metacloak \cite{metacloak}: that utilizes a pool of suurogate diffusion models to generate model-agnostic unlearnable samples. Following standard setup in \cite{metacloak}, we present our results at the allowed perturbation radius ($\ell_\infty$-norm ball) of $10/255$. We also present our results by varying the ($\ell_\infty$-norm ball) from $4/255$ to $16/255$ with a step size of $1$ as an ablation study. We utilize standard data transformations including Gaussian filtering with a kernel size of 7, horizontal flipping with half probability, center cropping, and image resizing to 512x512 as mentioned in \cite{metacloak} during training.

\newcommand{\cmark}{\textcolor{green}{\ding{51}}}
\newcommand{\xmark}{\textcolor{red}{\ding{55}}}
\definecolor{darkgreen}{HTML}{EBFFEA}

\begingroup
% tighten space below two-column floats and inside caption
\setlength{\dbltextfloatsep}{4pt plus 1pt minus 1pt}
\setlength{\textfloatsep}{4pt plus 1pt minus 1pt}
\captionsetup[table]{aboveskip=2pt, belowskip=0pt}

\begin{table*}[t]
\centering
\setlength\tabcolsep{4pt}  % Reduced column padding
\scalebox{0.725}{  % Adjust scale as needed
\begin{tabular}{l ccc ccc ccc ccccc ccccc}
\toprule
\multirow{2}{*}{Method} 
& \multicolumn{3}{c}{Imperceptibility (All)} 
& \multicolumn{3}{c}{CelebA-HQ (Face)} 
& \multicolumn{3}{c}{VGGFace2 (Face)} 
& \multicolumn{5}{c}{WikiArt (Non-Face)} 
& \multicolumn{5}{c}{DreamBooth (Non-Face)} \\
\cmidrule(lr){2-4} \cmidrule(lr){5-7} \cmidrule(lr){8-10} \cmidrule(lr){11-15} \cmidrule(lr){16-20}
& PSNR $\uparrow$ & SSIM $\uparrow$ & FID $\downarrow$ 
& ISM $\downarrow$ & BRIS. $\uparrow$ & FDFR $\uparrow$  % CelebA
& ISM $\downarrow$ & BRIS. $\uparrow$ & FDFR $\uparrow$  % VGGFace2
& SDS $\downarrow$ & ISM $\downarrow$ & BRIS. $\uparrow$ & IQAC $\downarrow$ & LIQE $\downarrow$  % WikiArt
& SDS $\downarrow$ & ISM $\downarrow$ & BRIS. $\uparrow$ & IQAC $\downarrow$ & LIQE $\downarrow$ \\ % DreamBooth
\midrule
AntiDB \cite{antidb} & \bignum{32.74} & \bignum{0.86} & \bignum{48.45} 
& \bignum{0.82} & \bignum{20.20} & \bignum{48.39}  % CelebA
& \bignum{0.84} & \bignum{25.29} & \bignum{40.21}  % VGGFace2
& \bignum{0.73} & \bignum{0.81} & \bignum{26.92} & \bignum{0.71} & \bignum{0.87}  % WikiArt
& \bignum{0.75} & \bignum{0.83} & \bignum{23.58} & \bignum{0.32} & \bignum{0.86} \\ % DreamBooth

Mist \cite{advdm} & \bignum{32.75} & \bignum{0.86} & \bignum{48.32} 
& \bignum{0.84} & \bignum{21.27} & \bignum{48.34}  % CelebA
& \bignum{0.86} & \bignum{26.93} & \bignum{42.40}  % VGGFace2
& \bignum{0.64} & \bignum{0.84} & \bignum{26.29} & \bignum{0.67} & \bignum{0.79}  % WikiArt
& \bignum{0.68} & \bignum{0.82} & \bignum{24.29} & \bignum{0.24} & \bignum{0.88} \\ % DreamBooth

AntiDB + SimAC \cite{simac} & \bignum{32.65} & \bignum{0.86} & \bignum{35.54} 
& \bignum{0.79} & \bignum{28.47} & \bignum{55.30}  % CelebA
& \bignum{0.71} & \bignum{29.75} & \bignum{64.27}  % VGGFace2
& \bignum{0.68} & \bignum{0.74} & \bignum{25.69} & \bignum{0.64} & \bignum{0.62}  % WikiArt
& \bignum{0.60} & \bignum{0.73} & \bignum{25.23} & \bignum{0.20} & \bignum{0.62} \\ % DreamBooth

MetaCloak \cite{metacloak} & \bignum{32.15} & \bignum{0.86} & \bignum{16.93} 
& \bignum{0.78} & \bignum{26.81} & \bignum{56.32}  % CelebA
& \bignum{\textbf{0.64}} & \bignum{30.37} & \bignum{65.93}  % VGGFace2
& \bignum{0.52} & \bignum{0.60} & \bignum{25.49} & \bignum{0.43} & \bignum{0.61}  % WikiArt
& \bignum{0.62} & \bignum{0.71} & \bignum{27.12} & \bignum{{-0.15}} & \bignum{{0.64}} \\ % DreamBooth

\rowcolor{darkgreen} Ours 
& \bignum{\textbf{42.27}} & \bignum{\textbf{0.95}} & \bignum{\textbf{16.03}} 
& \bignum{\textbf{0.58}} & \bignum{\textbf{38.61}} & \bignum{\textbf{82.39}}  % CelebA
& \bignum{0.60} & \bignum{\textbf{41.39}} & \bignum{\textbf{86.90}}  % VGGFace2
& \bignum{\textbf{0.48}} & \bignum{\textbf{0.52}} & \bignum{\textbf{36.26}} & \bignum{\textbf{0.37}} & \bignum{\textbf{0.46}}  % WikiArt
& \bignum{\textbf{0.49}} & \bignum{\textbf{0.53}} & \bignum{\textbf{37.29}} & \bignum{\textbf{-0.30}} & \bignum{\textbf{0.44}} \\ % DreamBooth
\bottomrule
\end{tabular}
}
\vspace{0.2em}
\caption{Quantitative results comparison to baseline methods: \textmd{\textit{Metrics: (1) Imperceptibility (PSNR, SSIM, FID), (2) Face-specific metrics (ISM, BRIS, FDFR), (3) Non-Face metrics (SDS, ISM, BRIS, IQAC, LIQE). The metrics presented utilize $10/255$ as the maximum budget, with $15$ steps of DiffPure purification after Unlearnable Sample Generation and TI training with the prompt ``A photo of an sks person/object'' for personalization. We generate $16$ images per prompt post personalization to report personalization metrics. We notice from the above table that our method shows significant improvements in imperceptible perturbation at the image-level while maintaining enhanced counter personalization metrics.}}}
\label{tab:main_table}
% \vspace{-0.1in}
\end{table*}
\endgroup
\vspace{-0pt}

After generation of unlearnable samples $\{\bar{x}^{\text{ul}}_0\}$, we apply DiffPure attack on generated unlearnable samples. We then leverage two prompts, ``a photo of sks person'' and ``a DSLR portrait of sks person'' to train Textual Inversion personalization. We generate \textit{16} images per prompt using these trained TI pipelines with $50$ inference steps. We present our results for countering personalization from these generated personalized images. For imperceptibility metrics, we utilize a frozen TI pipeline with the same random seed.

\begingroup
% tighten space below two-column floats and inside caption
\setlength{\dbltextfloatsep}{4pt plus 1pt minus 1pt}
\setlength{\textfloatsep}{4pt plus 1pt minus 1pt}
\captionsetup[figure]{aboveskip=2pt, belowskip=-8pt}
\begin{figure*}[t]
    \centering
    \includegraphics[width=\linewidth]{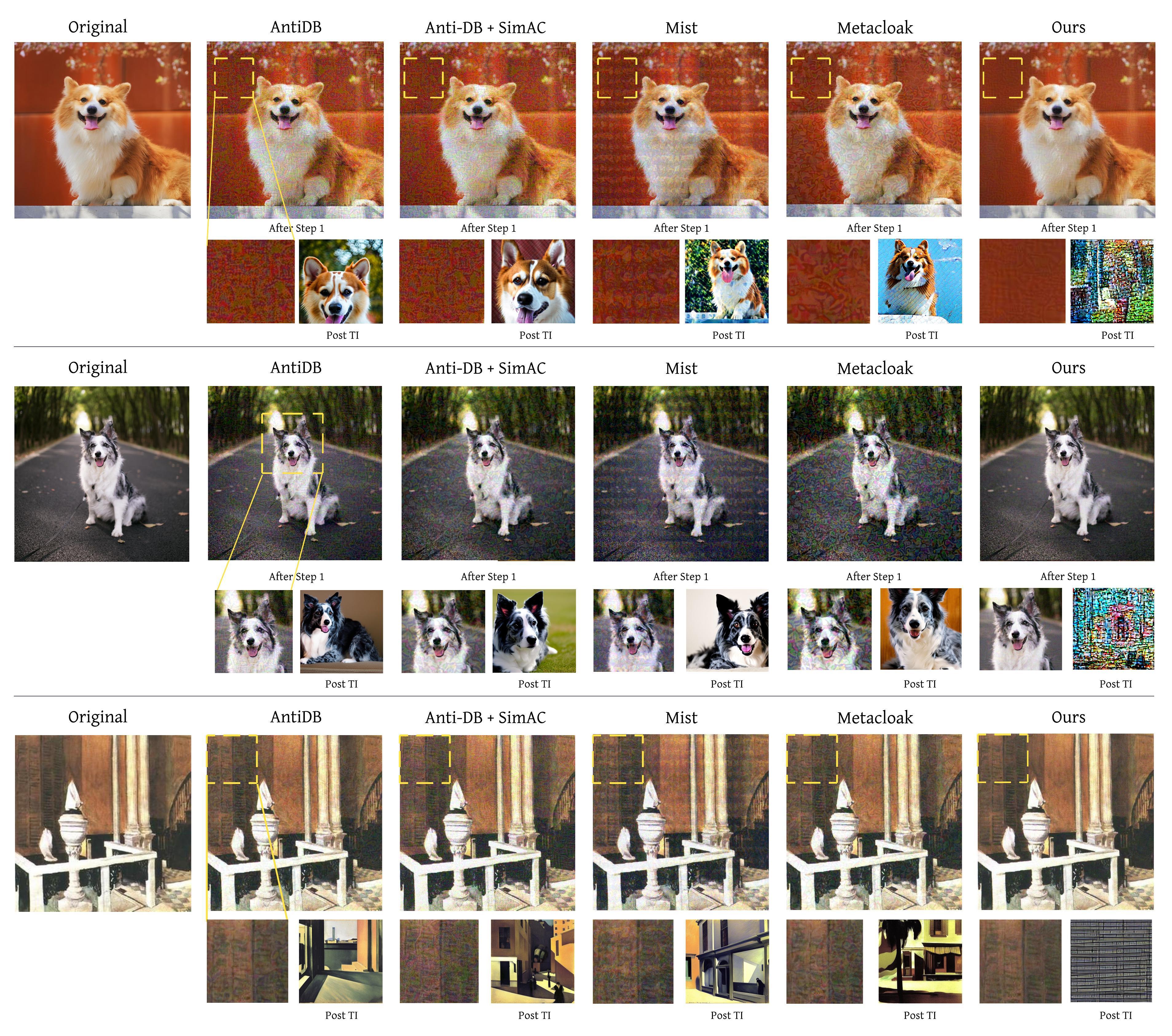}
    \vspace{-6mm}
    \caption{Qualitative Results Comparison to Baseline Methods: \textmd{\textit{Three illustrative examples with a maximum budget of $10/255$ compared to baselines. 
    Each example contains (top row) unlearnable samples after Step 1 followed by (bottom left) enlarged random region in the image to demonstrate difference in image perturbation, and (bottom right) Personalization result using TI after $15$ steps of DiffPure purification. We see that existing methods that rely on pixel-level perturbations add strong, visible noise/artifacts to the image without providing  identity protection. }} \textit{Best viewed on full-screen with zoomed view for clear difference in perturbation.}}
    \label{fig:main_qual}
\end{figure*}
\endgroup
\vspace{-6pt}

\subsection{Results on Unlearnable Sample Generation}

\cref{tab:main_table} presents the performance of our method compared to the baseline methods for unlearnable sample generation. We present all our metrics with DiffPure as our default attack after step 1. For imperceptibility, we average the results for PSNR, SSIM, and FID across four datasets. We see a significant improvement of over $8 \text{ dB}$ in PSNR compared to the best performing baseline \cite{advdm}. Our method perturbs the start of the denoising trajectory to perform unlearnable sample generation and achieves the best imperceptibility of unlearnable perturbation across all three metrics compared to all baseline methods. To test the performance of countering personalization, we utilize ``a photo of an sks person'' for Face datasets and ``a photo of an sks object'' for Non-Face datasets. Our method performs latent-level perturbation to generate unlearnable samples in the presence of a pre-trained diffusion based personalization method. This denoising-aware scenario of our method is expected to achieve enhanced robustness to denoising based adversarial attack such as DiffPure (strong adversary). For the face datasets, we see significant improvement of FDFR by over $20 \%$, and reduction in ISM from $0.78$ (best baseline method) to $0.582$. 

These metrics demonstrate the ability of our method to ensure that the identity in Face datasets is not maintained post personalization. We notice the same trend on Non-Face datasets where our method consistency outperforms several baselines with a significant increase of $10$ points on the BRISQUE metric both on WikiArt and DreamBooth datasets.

Qualitative results in \cref{fig:main_qual} demonstrate the effectiveness of our method in imperceptibility and robustness. 
We see that all the existing baselines add a very noticeable adversarial perturbation on the original image. We enlarge regions in the image for comparison with our method in the second row of each qualitative example. 
We also present the result of personalization after performing 15 DiffPure steps of purification. 
We notice that all the baselines fail to protect the identity of the person (or the object or the style) in the image, while our method ensures failure in personalization. 
Additional results from DreamBooth, WikiArt and VGGFace2, Celeba-HQ datasets are in the supplement.

\vspace{-0.1in}

\subsection{Ability to Transfer Unlearnable Sample Generation across Diffusion Pipelines}

\begin{table}[t]
\centering
\setlength{\tabcolsep}{3pt}
\begin{adjustbox}{scale=0.7}
\begin{tabular}{@{}ll*{10}{c}@{}}
\toprule
\multirow{2}{*}{Train} & \multirow{2}{*}{Test} & \multicolumn{5}{c}{``a photo of sks object''} & \multicolumn{5}{c}{``a dslr portrait of sks object''} \\
\cmidrule(lr){3-7} \cmidrule(lr){8-12}
 & & \raisebox{-0.5ex}{SDS $\downarrow$} & \raisebox{-0.5ex}{ISM $\downarrow$} & \raisebox{-0.5ex}{BRIS. $\uparrow$} & \raisebox{-0.5ex}{IQAC $\downarrow$} & \raisebox{-0.5ex}{LIQE $\downarrow$}
 & \raisebox{-0.5ex}{SDS $\downarrow$} & \raisebox{-0.5ex}{ISM $\downarrow$} & \raisebox{-0.5ex}{BRIS. $\uparrow$} & \raisebox{-0.5ex}{IQAC $\downarrow$} & \raisebox{-0.5ex}{LIQE $\downarrow$} \\
\midrule
v1.5 & v1.5 & \bignum{0.43} & \bignum{0.61} & \bignum{36.71} & \bignum{0.47} & \bignum{0.49} & \bignum{0.48} & \bignum{0.66} & \bignum{35.43} & \bignum{0.47} & \bignum{0.53} \\
v1.5 & v2.1 & \bignum{0.52} & \bignum{0.65} & \bignum{36.34} & \bignum{0.49} & \bignum{0.68} & \bignum{0.51} & \bignum{0.72} & \bignum{33.20} & \bignum{0.55} & \bignum{0.62} \\
v2.1 & v2.1 & \bignum{0.49} & \bignum{0.63} & \bignum{37.29} & \bignum{-0.33} & \bignum{0.44} & \bignum{0.52} & \bignum{0.64} & \bignum{34.54} & \bignum{0.39} & \bignum{0.51} \\
v2.1 & v1.5 & \bignum{0.32} & \bignum{0.54} & \bignum{40.86} & \bignum{-0.43} & \bignum{0.65} & \bignum{0.44} & \bignum{0.58} & \bignum{39.67} & \bignum{0.57} & \bignum{0.43} \\
\bottomrule
\end{tabular}
\end{adjustbox}
\vspace{0.2em}
\caption{Performance of different Stable Diffusion versions: \textmd{\textit{We test the generalization ability of our method by training and testing on different versions of Stable Diffusion (SD). 
The first prompt uses TI for personalization and the second prompt uses DB for personalization. 
We notice that our method can easily be plugged into different SD versions while countering personalization.}}}
\label{tab:sd_version_table}
\vspace{-0.4in}
\end{table}

Our method crafts unlearnable sample generation by shifting the latent at fully noised timestep $T$. Our method allows a seamless way to be plugged into various versions of Stable Diffusion pipelines. We test the generalization ability of our method by utilizing different versions of Stable Diffusion. We present our results in \cref{tab:sd_version_table} where we notice that our method maintains resistance to personalization when plugged into different SD versions. Recall from \cref{sec:training-conf} that our method can be trained in the presence of both Textual Inversion and DreamBooth personalization models as the pre-processing stage. In this section, we present the ability of our method to generalize between different versions of Stable Diffusion pipelines for TI and DB pre-processing stages in the form of two prompts for personalization. In table \cref{tab:sd_version_table} the first prompt (``a photo of sks object") represents TI personalization on unlearnable sample (post generation) and the second prompt (``a dslr portrait of sks object" represents DB personalization post generation.

We notice from \cref{tab:sd_version_table} that our method maintains consistent performance to counter personalization when trained on one SD version and tested on another. When trained and tested on the same SD version we expect performance to be maintained as presented in \cref{tab:main_table} as well. It is important to notice that when trained on SD v$1.5$ and tested on SD v$2.1$ our method maintains performance to ensure counter personalization. This performance improves when trained on SD v$2.1$ and evaluated on SD v$1.5$ demonstrating strong generalization ablity of our method across different versions of Stable Diffusion.

% Our method crafts unlearnable sample generation by shifting the latent at fully noised timestep $T$. In addition to crafting In-generation unlearnable samples, our method can be plugged into various versions of Stable Diffusion pipelines. We test the generalization ability of our method by utilizing different versions of Stable Diffusion. We utilize one Textual Inversion prompt and one DreamBooth prompt for personalization. We present our results in \cref{tab:sd_version_table} where we notice that our method maintains resistance to personalization when plugged into different SD versions.

% \SA{Does Table 2 gives generalization to both SD and different personalization? What it means for prompt one TI and prompt 2 DB?}

\vspace{-0.1in}

\subsection{Resistance to Purification Attacks}

\begin{figure}[!t]
    \centering
    \includegraphics[width=\linewidth]{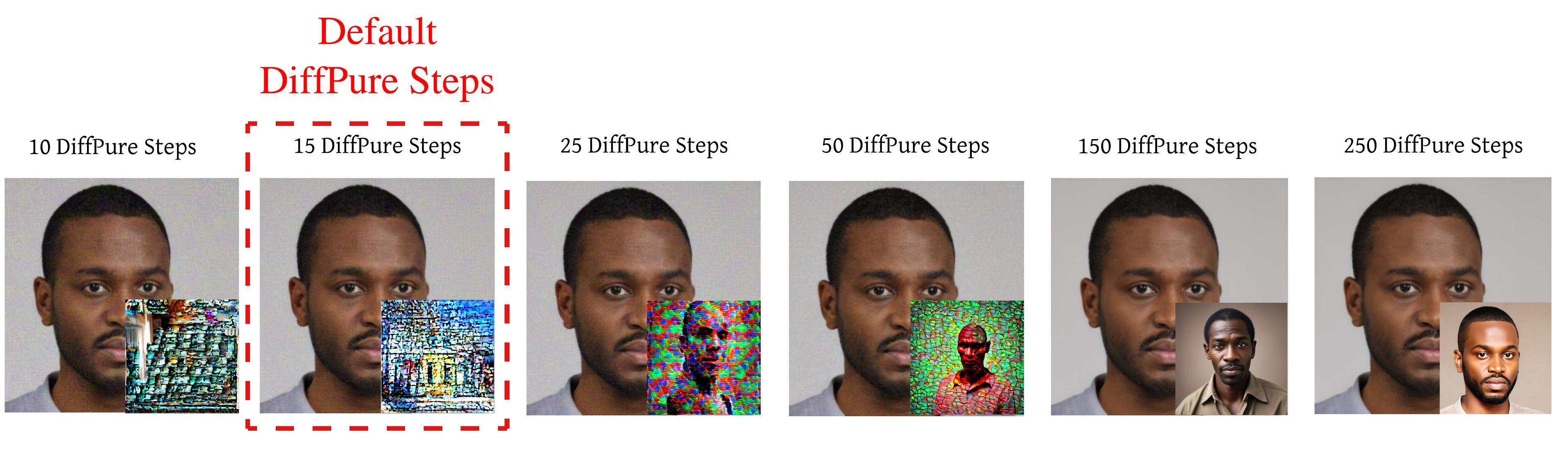}
    \vspace{-6mm}
    \caption{DiffPure Purification after Personalization: \textmd{\textit{We perform stress tests on the strength of DiffPure attack to demonstrate the resistance of our method to advanced adversarial based attacks. We see that our method is robust in preserving identity over $150$ DiffPure steps of purification. Only at around $250$ DiffPure steps, personalization takes place with the right identity.}}}
    \label{fig:diffpure_qual}
    \vspace{-0.1in}
\end{figure}

Recall that our method performs latent-level perturbation to generate unlearnable samples. As depicted in \cref{tab:main_table}, and \cref{fig:main_qual}, our method demonstrates enhanced robustness to strong adversarial DiffPure attack. We test the extent of robustness (of our method) to DiffPure purification attack and present the results in \cref{fig:diffpure_qual}. After Unlearnable Sample Generation, we pass the unlearnable image into DiffPure purification pipeline to retrieve a reconstructed samples. With these reconstructed samples, we perform personalization using TI and DB. We present the results at the end of personalization for $10, 25, 50, 150, \text{ and } 250$ DiffPure steps. We notice from \cref{fig:diffpure_qual} that as the number of steps in DiffPure increases, the adversarial perturbation in the images gets lower and lower, while the personalization fails successfully until over 150 DiffPure steps. This demonstrates the resistance of our method to a strong Diffusion based purification method. We chose DiffPure as the default attack as it peforms diffusion-based purification via reconstruction which is a powerful, specific attack and resistance to such powerful attack is desirable for unlearnable sample generation.

\vspace{-0.1in}

\subsection{Ablation Studies}

\paragraph{Ablation on Budget for Unlearnable Sample Generation} We present both qualitative \cref{fig:budget_qual} and quantitative \cref{tab:pert_strength_table} with various budget allowed during Step 1. We range budgets from $4/255$ to $32/255$ where more budget results in strong adversarial patch in the unlearnable sample that strengthens unlearnable behaviour v/s low budget (such as $4/255$) where achieving unlearnability might not be possible. In \cref{fig:budget_qual} we see this trend demonstrated on two samples. \cref{tab:pert_strength_table} presents quantitative metrics across $4$ prompts that use TI training for personalization. We see that as the budget for perturbation increases Subject Detection Score, ISM, IQAC, and LIQE consistently decrease, while BRISQUE increases simultaneously. In the bottom row of the table, we test personalization with two new prompts after TI and we notice the same consistent trend with countering personalization and the budget increases. From our empirical study, we identify an optimal budget $10/255$ that ensures imperceptibility and strong unlearnable behaviour and we utilize $10/255$ as the default budget to report our results.

\begin{table}[t]
\centering
\setlength{\tabcolsep}{3pt}
\begin{adjustbox}{scale=0.73}
\begin{tabular}{@{}l*{10}{c}@{}}
\toprule
\multirow{2}{*}{$\eta$} & \multicolumn{5}{c}{``a photo of sks object'} & \multicolumn{5}{c}{``a dslr portrait of sks object''} \\
\cmidrule(lr){2-6} \cmidrule(lr){7-11}
 & \raisebox{-0.5ex}{SDS $\downarrow$} & \raisebox{-0.5ex}{ISM $\downarrow$} & \raisebox{-0.5ex}{BRIS. $\uparrow$} & \raisebox{-0.5ex}{IQAC $\downarrow$} & \raisebox{-0.5ex}{LIQE $\downarrow$}
  & \raisebox{-0.5ex}{SDS $\downarrow$} & \raisebox{-0.5ex}{ISM $\downarrow$} & \raisebox{-0.5ex}{BRIS. $\uparrow$} & \raisebox{-0.5ex}{IQAC $\downarrow$} & \raisebox{-0.5ex}{LIQE $\downarrow$} \\
\midrule
4/255 & \bignum{0.53} & \bignum{0.70} & \bignum{35.79} & \bignum{-0.2} & \bignum{0.52} & \bignum{0.51} & \bignum{0.72} & \bignum{33.12} & \bignum{0.41} & \bignum{0.659} \\
8/255 & \bignum{0.49} & \bignum{0.63} & \bignum{37.29} & \bignum{-0.33} & \bignum{0.44} & \bignum{0.52} & \bignum{0.64} & \bignum{34.54} & \bignum{0.39} & \bignum{0.51} \\
12/255 & \bignum{0.35} & \bignum{0.58} & \bignum{38.59} & \bignum{-0.35} & \bignum{0.38} & \bignum{0.48} & \bignum{0.52} & \bignum{36.26} & \bignum{0.368} & \bignum{0.46} \\
32/255 & \bignum{0.28} & \bignum{0.47} & \bignum{42.96} & \bignum{-0.51} & \bignum{0.23} & \bignum{0.31} & \bignum{0.38} & \bignum{39.54} & \bignum{0.25} & \bignum{0.32} \\

\midrule[0.8pt]
\multicolumn{11}{@{}l}{} \\[-10pt]
\multirow{2}{*}{$\eta$} & \multicolumn{5}{c}{``looking at a mirror''} & \multicolumn{5}{c}{``in front of eiffel tower''} \\
\cmidrule(lr){2-6} \cmidrule(lr){7-11}
 & \raisebox{-0.5ex}{SDS $\downarrow$} & \raisebox{-0.5ex}{ISM $\uparrow$} & \raisebox{-0.5ex}{BRIS. $\uparrow$} & \raisebox{-0.5ex}{IQAC $\downarrow$} & \raisebox{-0.5ex}{LIQE $\uparrow$}
 & \raisebox{-0.5ex}{SDS $\downarrow$} & \raisebox{-0.5ex}{ISM $\uparrow$} & \raisebox{-0.5ex}{BRIS. $\uparrow$} & \raisebox{-0.5ex}{IQAC $\downarrow$} & \raisebox{-0.5ex}{LIQE $\uparrow$} \\
\midrule
4/255 & \bignum{0.51} & \bignum{0.72} & \bignum{33.12} & \bignum{0.59} & \bignum{0.63} & \bignum{0.52} & \bignum{0.72} & \bignum{33.17} & \bignum{0.51} & \bignum{0.60} \\
8/255 & \bignum{0.43} & \bignum{0.61} & \bignum{36.71} & \bignum{0.47} & \bignum{0.49} & \bignum{0.48} & \bignum{0.66} & \bignum{35.43} & \bignum{0.47} & \bignum{0.53} \\
12/255 & \bignum{0.38} & \bignum{0.55} & \bignum{37.32} & \bignum{0.33} & \bignum{0.42} & \bignum{0.42} & \bignum{0.60} & \bignum{38.35} & \bignum{0.41} & \bignum{0.48} \\
32/255 & \bignum{0.30} & \bignum{0.41} & \bignum{39.97} & \bignum{0.21} & \bignum{0.31} & \bignum{0.31} & \bignum{0.52} & \bignum{41.53} & \bignum{0.32} & \bignum{0.39} \\
\bottomrule
\end{tabular}
\end{adjustbox}
\vspace{0.1in}
\caption{Strength of Perturbation v/s Personalization metrics: \textmd{\textit{We study the trend between budget of perturbation v/s personalization metrics of unlearnable samples.}}}
\label{tab:pert_strength_table}
\vspace{-0.3in}
\end{table}

\begin{figure}[!t]
    \centering
    \includegraphics[width=\linewidth]{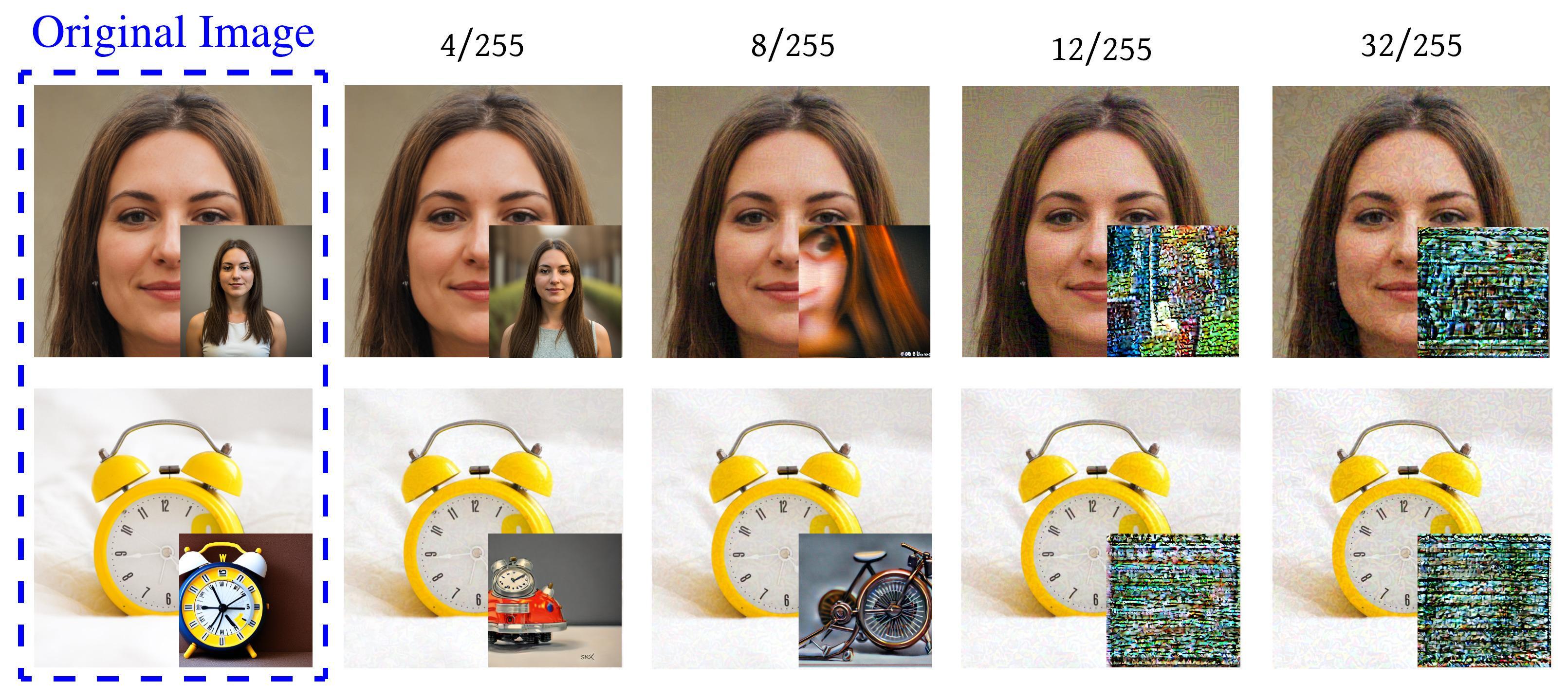}
    \vspace{-6mm}
    \caption{Budget of Perturbation v/s Personalization metrics: \textmd{\textit{We notice that as the budget increases, the adversarial perturbation is more significant but the unlearnable image becomes unusable (by the user as its significantly corrupted). We identify an optimal budget $10/255$ that we utilize for reporting our main results.}}}
    \label{fig:budget_qual}
    \vspace{-0.2in}
\end{figure}

\paragraph{\textbf{Ablation on Number of Denoising Steps for $\bar{z}^{\text{ul}}_0$ generation}:} The update at timestep $T$ during the Forward Diffusion process followed by denoising UNet during the Reverse Diffusion process to generate $z'_0$ have inherently oppposing goals where the second UNet is trying to pull the latent $z'_0$ toward original $z_0$ while the Unlearning UNet's aim to bring the trajectory shift to $z'_0$ to ensure unlearnable behaviour. We perform various empirical studies and test out different Reverse Diffusion processes to bring balance unlearnable ability v/s latent reconstruction. We utilize 1-step Diffusion models \cite{onestep} to perform denoising using the denoising UNet. We notice a trade-off between number of steps of reverse diffusion with the denoising UNet with 1-step update of the unlearning UNet. The trade-off between number of steps of denoising $\Phi^{\text{denoise}}$ and the scope to bring imperceptible unlearnable behaviour is also depicted in \cref{fig:method_motivation}. We notice that with 1-step denoising, the unlearning UNet dominates while hurting image quality, while significantly large number of denoising steps (>220) hurts counter-personalization. We balance this trade-off by finding an optimal number of denoising timesteps to be around $(k = 4)$ that provide the best benefits imperceptibility, robust unlearnable behaviour, and significant reduction in inference time for unlearnable image generation.

\begin{figure}[!t]
    \centering
    \includegraphics[width=\linewidth]{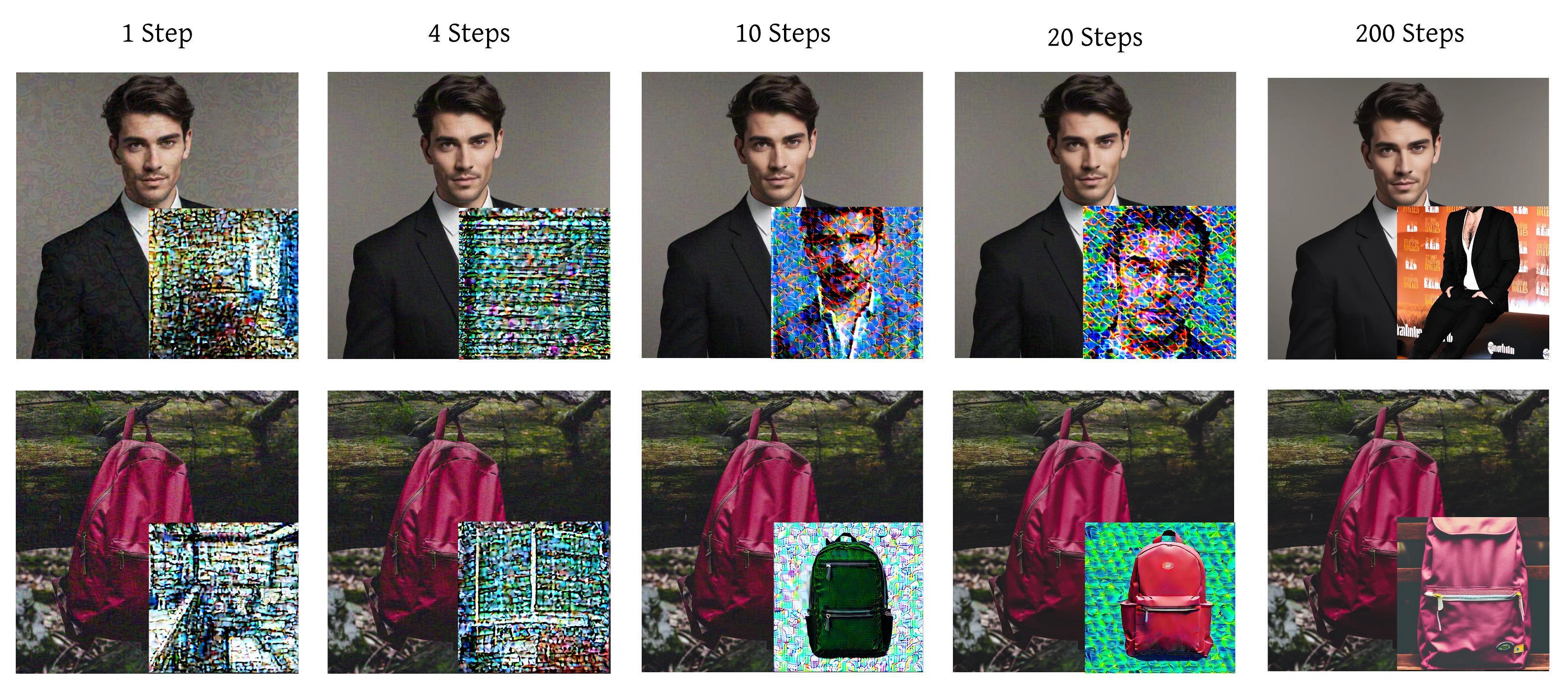}
    \vspace{-6mm}
    \caption{Number of Denoising steps v/s Personalization: \textmd{\textit{We identify a trade-off between number of denoising steps for unlearnable sample generation and countering personalization performance. We find an optimal timestep that preserves imperceptibility, robust unlearnable behaviour, and fast inference time.}}}
    \label{fig:denoising_qual}
    \vspace{-0.22in}
\end{figure}

\paragraph{\textbf{Computational overhead details}} Our latent perturbation module is a UNet consisting of stacked convolutional and transposed convolutional layers with ReLU/LeakyReLU activations and skip connections, totaling approximately 145K parameters. The architecture adds minimal computational overhead. It achieves inference times of $<0.2$ seconds per image and completes training in approximately $8 - 12$ hours across all four datasets. This design enables practical and scalable deployment without compromising unlearning performance.

% \vspace{-0.1in}

% \input{tables/strength_vs_qual}

\section{Conclusions}

In this work, we introduced a novel perturbation strategy that operates directly in the latent space of diffusion models to generate high-quality unlearnable examples. By shifting the denoising trajectory through a UNet-based perturbation model our method integrates unlearnable behavior into the framework of Latent Diffusion Models (LDMs) without significantly compromising visual quality. Unlike prior approaches, our method adds minimal perturbation when crafting unlearnable samples. Extensive experiments across four benchmark datasets, show that our approach offers strong robustness against advanced purification techniques such as DiffPure, while maintaining imperceptibility to the human eye. Quantitatively, we achieve notable gains in perceptual quality (PSNR, SSIM, FID improvements of $\sim8\%-10\%$) and robustness ($\sim10\%$ average increase across five adversarial scenarios).

% \vspace{-0.1in}

\paragraph{\textbf{Acknowledgments}} Prof. Lokhande thanks support provided by University at Buffalo Startup funds, Adobe Research Gift and internal funding from the University at Buffalo’s Research and Economic Development office. Dr. Tejas Gokhale was supported by UMBC’s Strategic Award for Research Transitions (START).

% These results underscore the practicality and generalizability of our approach as a scalable, plug-and-play defense against unauthorized model personalization and inversion, marking a significant step forward in safeguarding visual data privacy.

\bibliographystyle{acm}
\balance
\bibliography{references}

\newpage
\title{Latent Diffusion Unlearning: Protecting against Unauthorized Personalization through Trajectory Shifted Perturbations}

\section*{Supplementary Material}
\addcontentsline{toc}{section}{Supplementary Material} % optional, for TOC

\appendix
\renewcommand{\thesection}{A\arabic{section}} % Sections become A1, A2, ...
\setcounter{section}{0} % restart numbering

\section{Additional Details on Evaluation Metrics}

To comprehensively assess the effectiveness of our proposed unlearnable sample generation method, we utilize a wide range of quantitative metrics that evaluate three critical properties: (i) imperceptibility between original and perturbed samples, (ii) resistance to personalization techniques, and (iii) robustness against purification-based adversarial attacks. Below, we provide detailed descriptions and references for each evaluation metric used in our experiments.

\subsubsection{Imperceptibility Evaluation Metrics}

To ensure that the generated unlearnable sample $\bar{x}_0^{\text{ul}}$ remains visually indistinguishable from the original input $x_0$, we evaluate imperceptibility using the following widely-used image quality metrics:

\begin{itemize}
\item \textbf{Peak Signal-to-Noise Ratio (PSNR):}
PSNR measures the pixel-level fidelity between two images and is computed as a logarithmic ratio between the maximum possible pixel intensity and the Mean Squared Error (MSE) between the original and perturbed images. Higher PSNR values imply better visual similarity. We follow the standard definition of PSNR as in \cite{psnr}.

\item \textbf{Structural Similarity Index Measure (SSIM):}  
SSIM is a perceptual metric that quantifies image quality degradation based on changes in structural information, contrast, and luminance. It is bounded in $[0, 1]$, where higher values indicate better preservation of structural similarity. We use the multiscale SSIM (MS-SSIM) variant where appropriate, as introduced in \cite{wang2003multiscale, wang2004image}.

\item \textbf{Fréchet Inception Distance (FID):}  
FID evaluates the perceptual similarity between two sets of images using deep feature distributions extracted from a pretrained Inception network. It compares the mean and covariance of features for $x_0$ and $\bar{x}_0^{\text{ul}}$ and is particularly sensitive to both low-level visual quality and high-level semantic fidelity. Lower FID scores indicate greater similarity. We use the official FID implementation from \cite{heusel2017gans}.

\end{itemize}

These metrics jointly ensure that our unlearnable images are imperceptibly close to the originals in both low-level and high-level visual domains.

\subsubsection{Countering Personalization Metrics}

To verify that our generated unlearnable samples $\bar{x}_0^{\text{ul}}$ effectively degrade downstream model personalization, we evaluate them within a full personalization pipeline using Textual Inversion (TI) \cite{gal2022image}. We apply the following metrics to quantify the extent to which personalization fails when trained on unlearnable images:

\begin{itemize}
\item \textbf{Identity Score Matching (ISM):}
ISM computes the cosine similarity between identity embeddings of original and personalized images using a face recognition or identity classifier. A lower ISM score indicates greater deviation from the target identity, thereby reflecting stronger unlearnability.

\item \textbf{Subject Detection Score (SDS):}  
SDS measures the ability of a model to localize and identify subjects from a personalized model trained on $\bar{x}_0^{\text{ul}}$. It is derived from subject classification or segmentation performance metrics. Degraded SDS reflects effective suppression of personalization.

\item \textbf{Blind/Referenceless Image Spatial Quality Evaluator (BRISQUE):}  
BRISQUE is a no-reference quality assessment metric that estimates image degradation using natural scene statistics \cite{mittal2012no}. It is useful for quantifying quality drops introduced by imperceptible perturbations in the absence of ground truth.

\item \textbf{Face Detection Failure Rate (FDFR):}  
For face-centric datasets, we use FDFR to measure the percentage of $\bar{x}_0^{\text{ul}}$ images for which standard face detectors (e.g., MTCNN or DLib) fail to detect any facial region. Higher FDFR values imply successful suppression of facial features critical to personalization.

\item \textbf{CLIP-IQAC and LIQE:}  
For non-face datasets, we include CLIP-based Image Quality Assessment Contrastive Score (CLIP-IQAC) and Learned Image Quality Evaluator (LIQE) \cite{madhusudana2023image} to evaluate perceptual consistency and quality without ground truth. These models operate using contrastive and learned representations respectively and serve as complementary no-reference quality metrics.

\end{itemize}

\subsubsection{Robustness Evaluation under Purification Attacks}

Given the increasing prevalence of purification-based defenses in generative and personalization systems, we also evaluate the robustness of our method under such adversarial settings. Specifically, we consider the DiffPure defense \cite{nie2022diffpure}, a strong denoising-based purification attack that leverages diffusion models to remove adversarial perturbations.

We insert the DiffPure module between unlearnable sample generation and the personalization pipeline to assess whether $\bar{x}_0^{\text{ul}}$ still degrades performance after being denoised. Unlike pixel-space perturbation techniques that are often neutralized by such defenses, our method performs perturbation in the latent space and maintains its adversarial effectiveness even after purification. This is because latent perturbations strategically disrupt the diffusion trajectory, making the denoising process less effective in reversing them.

Robustness is again evaluated using ISM, BRISQUE, SDS, and FDFR/LIQE, after purification. Our method’s resilience to DiffPure underscores the importance of targeting latent diffusion pathways for robust unlearnable sample generation.

\section{Additional Details on Baselines Implementation}

To ensure a rigorous and meaningful comparison, we benchmark our method against a diverse set of state-of-the-art techniques designed to generate unlearnable examples that hinder text-to-image personalization. Below, we detail the implementation configurations for each baseline, adhering strictly to their original papers and available open-source repositories when applicable.

\begin{itemize}
\item \textbf{AntiDB \cite{antidb}:}
AntiDB introduces Fully-trained Surrogate Model Guidance (FSMG) and Alternating Surrogate and Perturbation Learning (ASPL), which together aim to craft perturbations that degrade DreamBooth-based personalization. For fair evaluation, we implement AntiDB with the official surrogate architecture and hyperparameter schedules. Perturbations are optimized using ASPL over 1,000 iterations with Adam optimizer (learning rate 0.005) within an $\ell_\infty$ constraint of $10/255$. We apply random affine transformations and intensity jittering during training, as prescribed in their paper.

\item \textbf{Mist (AdvDM) \cite{advdm}:}  
Mist leverages pre-trained diffusion models to generate adversarial examples targeted against Textual Inversion (TI). The method adds perturbations to pixel-space inputs during the forward diffusion process, using the diffusion gradients to direct the adversarial update. We implement Mist using Stable Diffusion v2.1 as the base model, applying a fixed guidance scale of 7.5. Perturbations are optimized via projected gradient descent over 500 steps within the allowed $\ell_\infty$-norm of $10/255$.

\item \textbf{AntiDB $+$ SimAC \cite{simac}:}  
SimAC introduces a greedy timestep selection heuristic that determines which timestep in the forward diffusion process is most susceptible to adversarial perturbation. The selected timestep is then perturbed, and reverse diffusion is applied to reconstruct the image. In combination with AntiDB, SimAC improves personalization resistance. Our implementation adheres to the greedy timestep search from the original paper, restricting perturbations to a single optimized timestep ($t^\ast$) in the diffusion trajectory. Training is performed using a cosine learning rate decay.

\item \textbf{MetaCloak \cite{metacloak}:}  
MetaCloak adopts a meta-learning strategy to generate model-agnostic unlearnable examples by training against an ensemble of surrogate diffusion models. We reproduce the MetaCloak setup using four surrogate networks (all variants of U-Net with different random seeds and frozen parameters). Meta-optimization is done using MAML with inner loop updates of 3 steps and an outer loop learning rate of 0.0005. Following the standard evaluation protocol from \cite{metacloak}, we restrict perturbations to lie within an $\ell_\infty$-norm ball of $10/255$. Additionally, we perform an ablation study where the allowed perturbation budget varies from $4/255$ to $16/255$ (step size 1), and we report trends in imperceptibility and personalization resistance. Training-time data augmentations include Gaussian filtering (kernel size = 7), horizontal flipping (probability = 0.5), center cropping, and resizing to $512 \times 512$, as in the original paper.

\end{itemize}

\vspace{1mm}
\noindent
\textbf{Post-processing and Evaluation Setup:}
Once the unlearnable image set $\{\bar{x}^{\text{ul}}_0\}$ is generated for each method, we evaluate its robustness to purification using the \textbf{DiffPure} defense \cite{nie2022diffpure}. DiffPure is inserted as an intermediate purification module prior to the personalization step. The purified $\bar{x}^{\text{ul}}_0$ samples are then passed into a Textual Inversion (TI) pipeline.

For personalization, we follow a standardized evaluation setup across all baselines:

\begin{itemize}
\item \textbf{Prompt Selection:} We use multiple prompts per subject, such as — “\textit{a photo of sks person}” and “\textit{a DSLR portrait of sks person}” — which are consistent with prior personalization benchmarks.
\item \textbf{TI Training:} For each method, TI is trained using $6$ unlearnable images with 1500 steps of fine-tuning, a learning rate of $5e^{-4}$, and a batch size of 1.
\item \textbf{Sampling Configuration:} After training, we generate 16 personalized outputs per prompt using the frozen TI model and 50 inference steps via DDIM sampling.
\item \textbf{Imperceptibility Metrics:} To ensure a fair assessment of visual similarity, imperceptibility metrics (PSNR, SSIM, FID) are computed using a frozen TI pipeline with a fixed seed. This guarantees consistency in image decoding across baselines.
\end{itemize}

This standardized pipeline enables direct comparison of imperceptibility, personalization degradation, and robustness across our latent perturbation method and all considered pixel-space baselines.

\section{Pre-Processing/Pre-Training Textual Inversion before Unlearnable Sample Generation}

Personalization methods including Textual Inversion (TI) and DreamBooth (DB). These models act as downstream personalization mechanisms which our perturbation generator $\rho$ is explicitly trained to counteract.

\subsubsection{Textual Inversion Token Pre-Training}

For each object or identity instance in the dataset, we pre-train a unique Textual Inversion token embedding following the original method in \cite{gal2022image}. Given a fixed prompt structure (e.g., “a photo of [$V_*$] person”), a pseudo-token [$V_*$] is optimized to capture identity-specific visual features in the embedding space of a frozen text encoder (e.g., CLIP). This embedding learns to steer the diffusion model toward personalized generations for that identity. Each identity in our dataset is thus associated with a distinct TI token.

The learned TI token embeddings $\tau_i$ serve as the personalization anchors which simulate real-world model customization using a small set of images. These embeddings are frozen and reused during unlearnable sample generation.

\subsection{Training Adversarial Latents Against Pre-Trained Personalization Models}

Once TI tokens are learned, we train our perturbation model $\rho$ to generate unlearnable latent samples $\bar{z}_0^{\text{ul}}$ by introducing adversarial noise at the noised latent $z_T$ stage. The goal is to produce perturbed reconstructions $\bar{x}_0^{\text{ul}}$ which, when used in combination with the pre-trained TI tokens, fail to personalize effectively.

\subsection{Token Embedding Shuffling for Robustness}

To further enhance robustness, we introduce TI token shuffling during unlearnable sample training. Instead of pairing an identity $i$ with its corresponding token $\tau_i$ in every training step, we randomly assign TI tokens from other identities $\tau_j$, $j \neq i$, as distractors during the personalization phase. This randomized pairing helps the model generalize against multiple personalization anchors and prevents overfitting to a specific identity-token binding.

This shuffling operation increases the diversity of personalization failure cases seen during training, thereby enhancing the adversarial generalization of our perturbation module $\rho$. It ensures that our generated unlearnable images do not just counter a single TI embedding, but maintain resistance across a distribution of potential personalization targets.

Although our primary experiments use TI due to its compactness and reproducibility, we also extend the same setup to DreamBooth \cite{Ruiz_2023_CVPR}, where a personalized fine-tuned model is trained for each identity. In this case, unlearnable samples are trained to degrade the quality of generated outputs from a DreamBooth-tuned model. The shuffling strategy is similarly employed by varying subject instance-to-model mappings during adversarial training.

In addition to ``a photo of sks person", and ``a portrait of sks person" prompts, we also test our method with different validation prompts as well. We mention the results of these prompts Table 3 of our main paper. The prompts ``looking at the mirror" and ``in front of eiffel tower" are not seen during training.

\section{Towards Continual Training for Unlearnable Sample Generation}

Most existing unlearnable sample generation methods assume a static setup, where adversarial perturbations are crafted after a personalization model (such as Textual Inversion or DreamBooth) has already been trained and fixed. However, such decoupled training pipelines are inherently limited, as they do not account for adaptive personalization techniques that continually improve their representations through incremental fine-tuning. We aim towards a method that enables continual training of unlearnable samples in the presence of active personalization.

\subsection{Joint Optimization Framework}

Continual learning for unlearning requires dual-objective training loop where the adversarial UNet $\rho$ (used to perturb the noisy latent $z_T$) is trained simultaneously with the personalization model (e.g., Textual Inversion).

Formally, for each image $x_0$ associated with identity $i$, we optimize:

\begin{itemize}
    \item The TI embedding $\tau_i$ to maximize fidelity between generated images and identity features.
    \item The adversarial UNet $\rho$ to minimize personalization performance when $\bar{x}_0^{\text{ul}}$ (the perturbed version of $x_0$) is used for training.
\end{itemize}

This min-max formulation is written as:

\begin{equation}
\min_{\rho} \max_{\tau_i} \; \mathcal{L}^{\text{TI}}(x_0, \tau_i; \rho) - \lambda \cdot \mathcal{L}_{\text{unlearn}}(\bar{x}_0^{\text{ul}}, \tau_i)
\end{equation}

Here, $\mathcal{L}^{\text{TI}}$ captures the personalization success, while $\mathcal{L}_{\text{unlearn}}$ captures personalization degradation due to the perturbed data.

\subsection{Stabilizing Dynamics}

Ensuring stable training under this adversarial setting is a crucial requirement. Gradual change in TI embedding parameters and Unet $\rho$ parameters must be maintained to avoid collapse:

\begin{itemize}
    \item TI embeddings remain meaningful, avoiding collapse.
    \item $\rho$ learns to craft perturbations that are resilient against increasingly optimized personalization embeddisngs.
\end{itemize}

\subsection{Benefits of Continual Training}

This continual training paradigm offers several key advantages:

\begin{itemize}
    \item Adaptivity to Personalization Drift: As the TI embeddings evolve, $\rho$ learns to counter them in real time, enabling robustness to evolving models.
    \item This setup better reflects real-world usage, where users may upload new training data, and personalization systems are incrementally updated.
\end{itemize}

\begin{figure*}[!h]
    \centering
    \includegraphics[width=\linewidth]{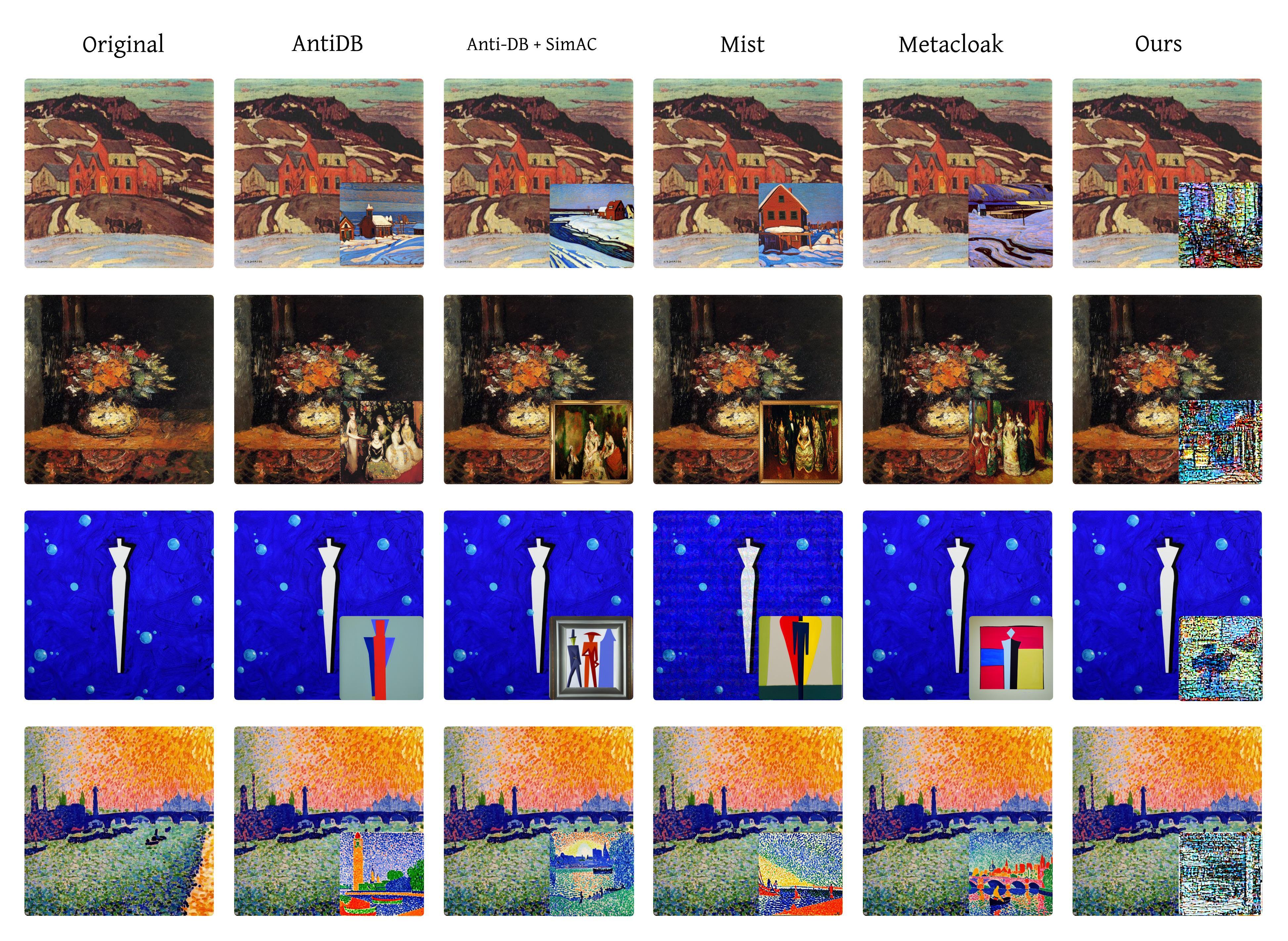}
    \vspace{-6mm}
    \caption{Additional Qualitative Results Comparison to Baseline Methods (WikiArt)}
    \label{fig:suppl_wiki}
\end{figure*}

\begin{figure*}[!h]
    \centering
    \includegraphics[width=\linewidth]{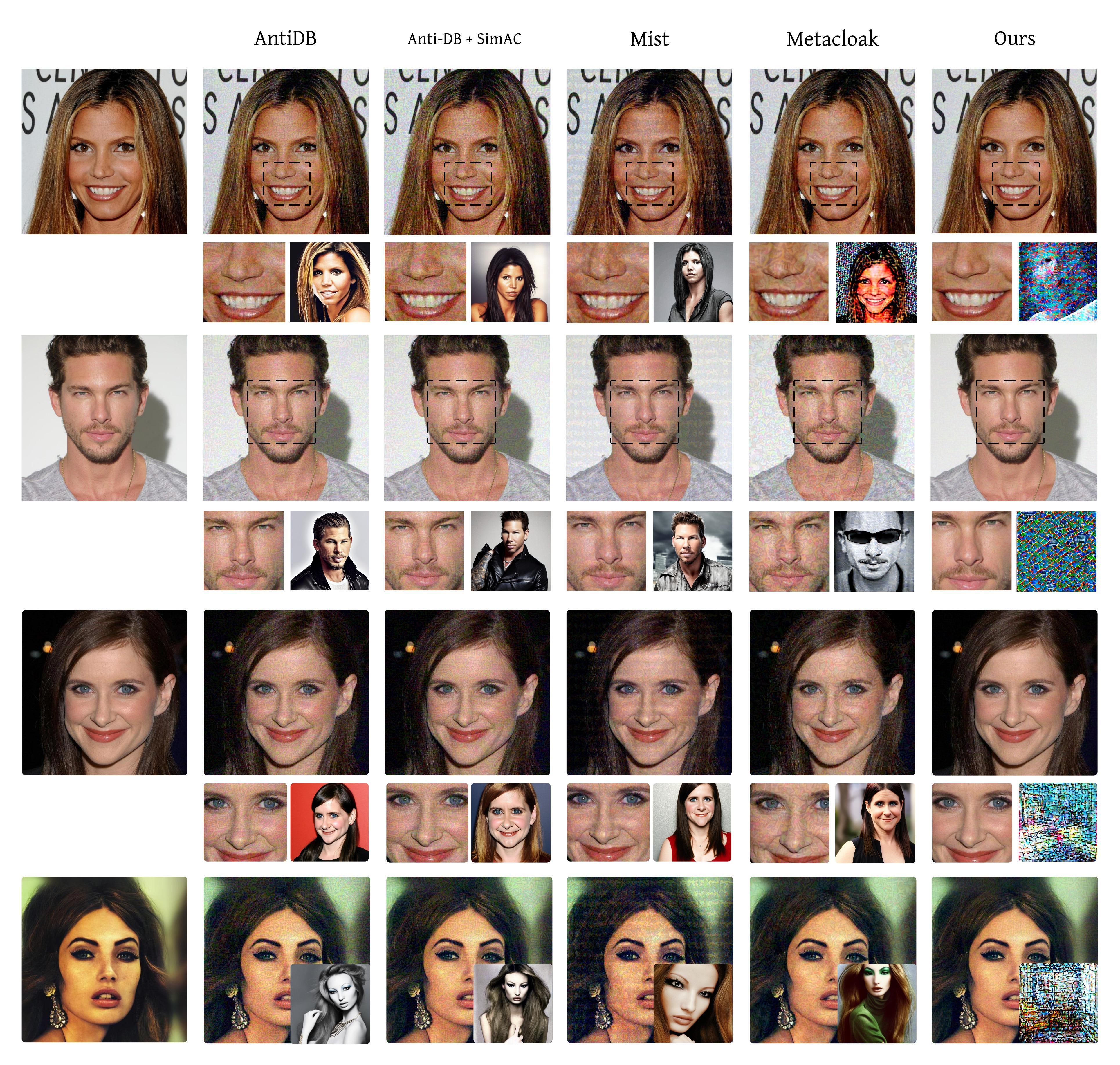}
    \vspace{-6mm}
    \caption{Additional Qualitative Results Comparison to Baseline Methods (Celeba-HQ)}
    \label{fig:suppl_celeba}
\end{figure*}

\begin{figure*}[!h]
    \centering
    \includegraphics[width=\linewidth]{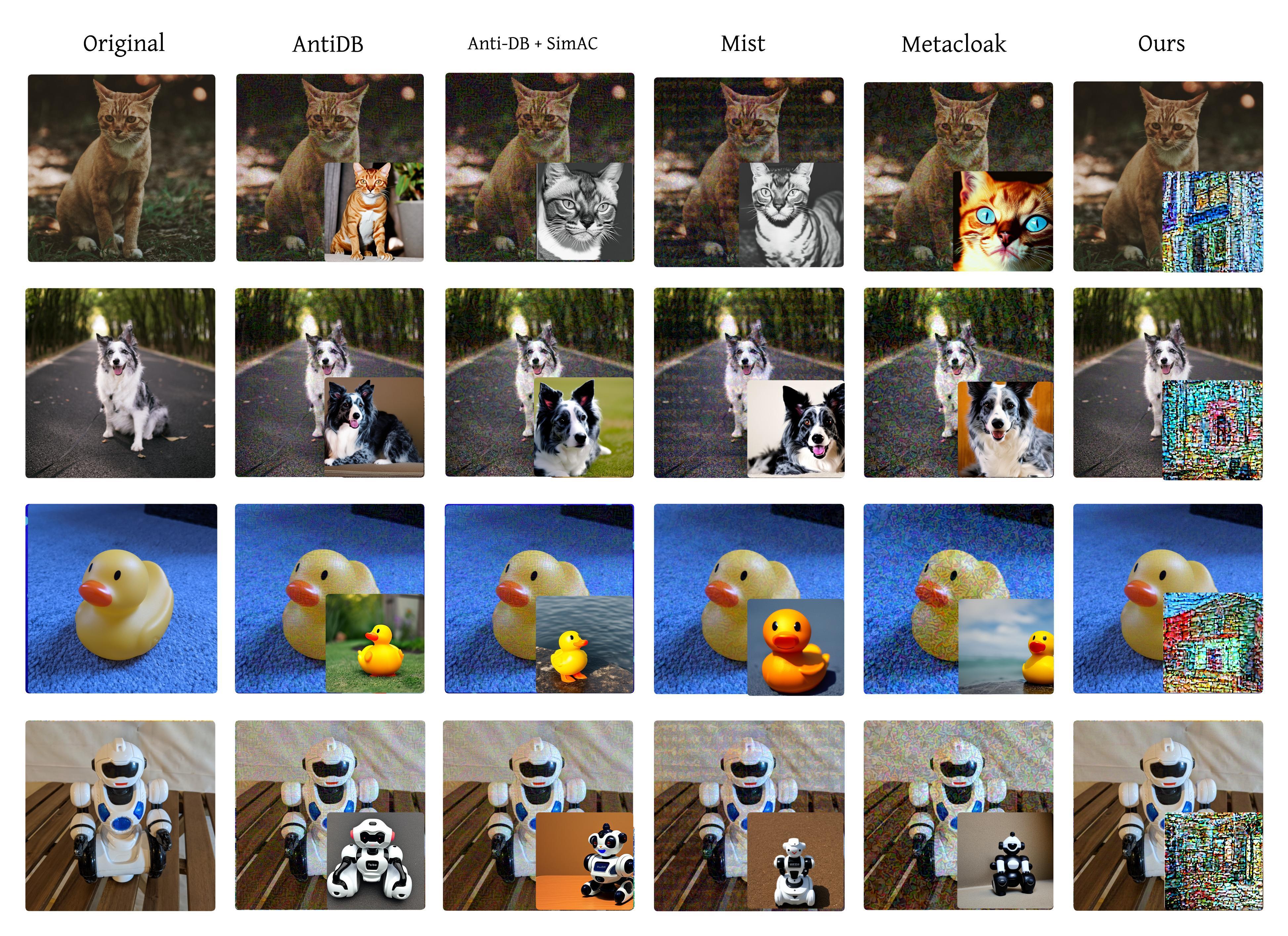}
    \vspace{-6mm}
    \caption{Additional Qualitative Results Comparison to Baseline Methods (DB Dataset)}
    \label{fig:suppl_db}
\end{figure*}

\begin{figure*}[!h]
    \centering
    \includegraphics[width=\linewidth]{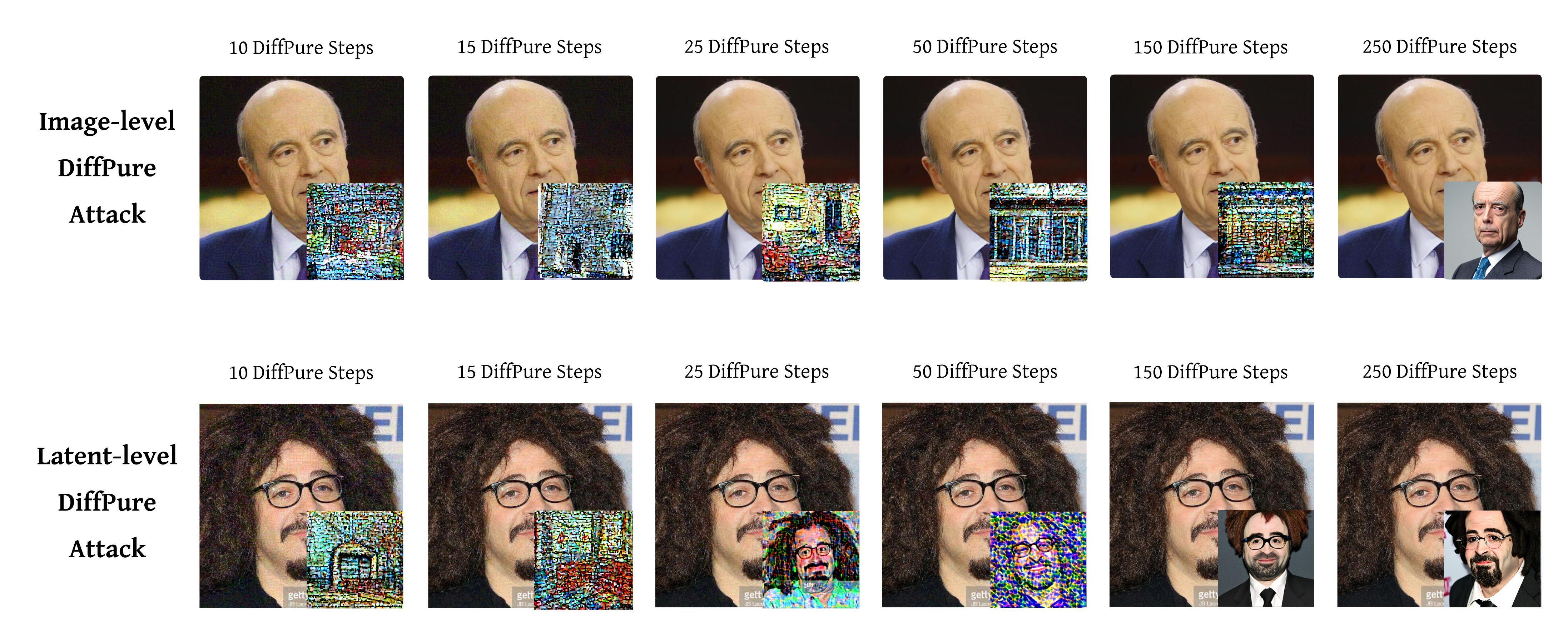}
    \vspace{-6mm}
    \caption{Robustness to Pixel-Level and Latent-Level DiffPure attack}
    \label{fig:suppl_diff}
\end{figure*}

\begin{figure*}[!h]
    \centering
    \includegraphics[width=0.6\linewidth]{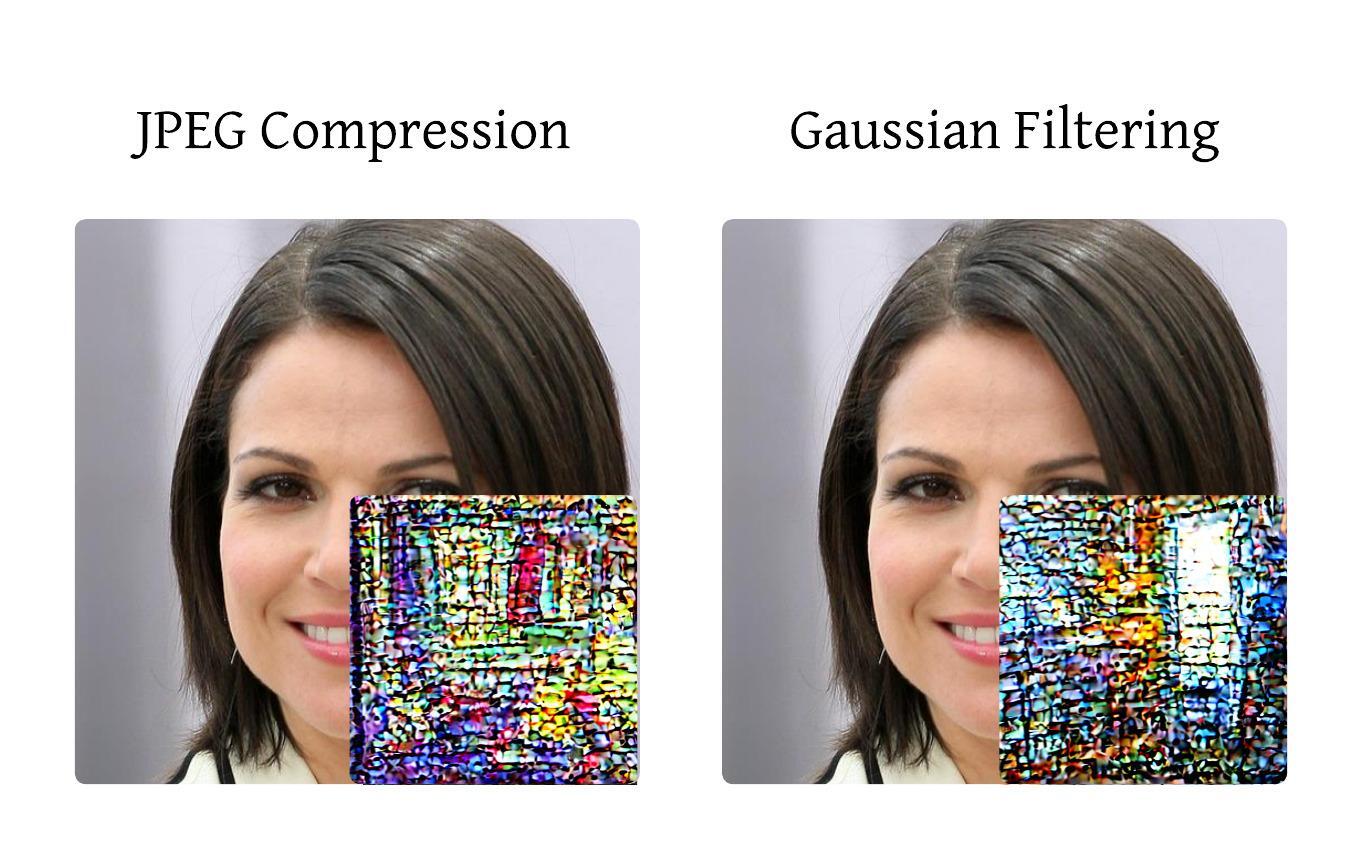}
    \vspace{-6mm}
    \caption{Robustness to JPEG Compression and Gaussian Filtering attack}
    \label{fig:attack_supp}
\end{figure*}

\end{document}